\pdfoutput=1
\documentclass[11pt]{article}
\usepackage[final]{acl}
\usepackage[table]{xcolor}
\usepackage{times,latexsym,fix-cm}
\DeclareFontShape{U}{lasy}{m}{n}{<->lasy10}{}
\usepackage[T1]{fontenc}
\usepackage[utf8]{inputenc}
\usepackage{microtype,booktabs,amsmath,amssymb,graphicx}
\usepackage{placeins}
\usepackage{float}
\floatstyle{ruled}
\newfloat{algorithm}{t}{loa}
\floatname{algorithm}{Algorithm}
\usepackage{tikz}
\usepackage[skins,breakable]{tcolorbox}
\usetikzlibrary{arrows.meta,positioning,calc,shapes.geometric}
\usepackage{url}
\usepackage{xurl}
\definecolor{inkblue}{RGB}{29,73,117}
\definecolor{teal}{RGB}{18,128,125}
\definecolor{mutedred}{RGB}{173,78,73}
\newtcolorbox{executioncase}[1]{enhanced,breakable,
 colback=black!1,colframe=black!48,boxrule=0.4pt,arc=2pt,
 left=7pt,right=7pt,top=8pt,bottom=6pt,
 before skip=9pt,after skip=3pt,
 fontupper=\small\raggedright,fonttitle=\small\bfseries,
 colbacktitle=black!60,coltitle=white,
 attach boxed title to top left={xshift=7pt,yshift=-2pt},
 minipage boxed title*=-24pt,halign title=flush left,
 boxed title style={boxrule=0pt,boxsep=0pt,arc=1.5pt,left=5pt,right=5pt,top=2pt,bottom=2pt},
 title={#1}}

\title{CompSkill: Compiling Execution Experience for Skill Retrieval}
\hypersetup{pdftitle={CompSkill: Compiling Execution Experience for Skill Retrieval},pdfauthor={Qi Shen, Ziyue Zhu, Yongqing Jiang, Zhen Xia, Li Shengjie, Shuai Zhao, Bo Cheng, Xiaobao Wu, Anh Tuan Luu, Haoran Luo}}
\author{Qi Shen$^{1}$\thanks{Equal contribution.}\quad Ziyue Zhu$^{1}$\footnotemark[1]\quad Yongqing Jiang$^{2,3}$\\
\bfseries Zhen Xia$^{1}$\quad Li Shengjie$^{1}$\quad Shuai Zhao$^{1}$\quad Bo Cheng$^{1}$\\
\bfseries Xiaobao Wu$^{4}$\quad Anh Tuan Luu$^{2}$\quad Haoran Luo$^{2}$\thanks{Corresponding author.}\\[4pt]
$^{1}$Beijing University of Posts and Telecommunications\quad $^{2}$Nanyang Technological University\\
$^{3}$Sichuan University\quad $^{4}$Shanghai Jiao Tong University}
\usepackage{tabularx,array,multirow}
\definecolor{ink}{HTML}{1E2430}\definecolor{accent}{HTML}{3F7F5F}
\definecolor{blue}{HTML}{1B3B6F}\definecolor{muted}{HTML}{5F6670}
\definecolor{brick}{HTML}{B5482F}\definecolor{amber}{HTML}{C8963E}
\definecolor{panel}{HTML}{E9EDF3}\definecolor{tint}{HTML}{EAF6F2}

\definecolor{hgpurple}{HTML}{7A4E7E}\definecolor{oxblood}{HTML}{9E2B25}

\newcommand{\na}{\textcolor{muted}{---}}
\newcommand{\yes}{\textcolor{accent}{\checkmark}}\newcommand{\no}{\textcolor{brick}{$\times$}}

\newcommand{\tbd}[1][]{\textcolor{muted}{\textsf{NR}}\ifx\relax#1\relax\else\textcolor{muted}{\textsf{[#1]}}\fi}

\newtheorem{proposition}{Proposition}
\renewcommand{\topfraction}{.92}\renewcommand{\dbltopfraction}{.92}\renewcommand{\textfraction}{.06}
\renewcommand{\floatpagefraction}{.88}\renewcommand{\dblfloatpagefraction}{.88}
\begin{document}
\makeatletter\clubpenalty=10000\@clubpenalty=\clubpenalty\widowpenalty=10000\makeatother
\maketitle
\suppressfloats[t]
\begin{abstract}
LLM agents increasingly rely on skill documents, yet most retrievers rank them by description alone and ignore what earlier executions revealed, even though a relevant skill can hurt execution.
We present CompSkill, which indexes skills with learned 512-d keys and compiles matched same-task comparisons into skill offsets and capped key residuals (KeyOpt), so experience changes which document the executor receives, never its text. Under a logistic outcome model with unknown task difficulty, matched comparisons cancel that difficulty and only discordant ones carry information; on support executions, predictions improve with each discordant task.
An experience hypergraph keeps each comparison with its executor and document versions, recalls missed skills and grounds a cited review.
Compact keys raise SkillRet NDCG@10 from 64.0 to 70.9 at one eighth of the dimension. With one document, CompSkill completes 256 and 260 of 332 ToolQA tasks with GPT-5.6-luna and DeepSeek-V4.1-Flash, respectively, more than every baseline, and stays on par with the strongest baselines on AppWorld and ALFWorld.
On 332 newly generated ToolQA tasks, it completes 256 with GPT-5.6-luna, significantly more than BM25, SkillRet-8B, Tool2Vec and RankGPT; over all 844 ToolQA tasks, it completes 24 more than the strongest baseline.
Selection matters beyond relevance: pairing the right guide with a same-database sibling instead of another database's guide costs 8 of 96 tasks.
Code, models and \mbox{support data are public.}{\hypersetup{linkcolor=black,urlcolor=black}\footnote{\href{https://github.com/satoshinji2992/CompSkill}{https://github.com/satoshinji2992/CompSkill}}}
\end{abstract}

\section{Introduction}
\label{sec:introduction}
\begin{figure}[t]
\centering
\includegraphics[width=\columnwidth]{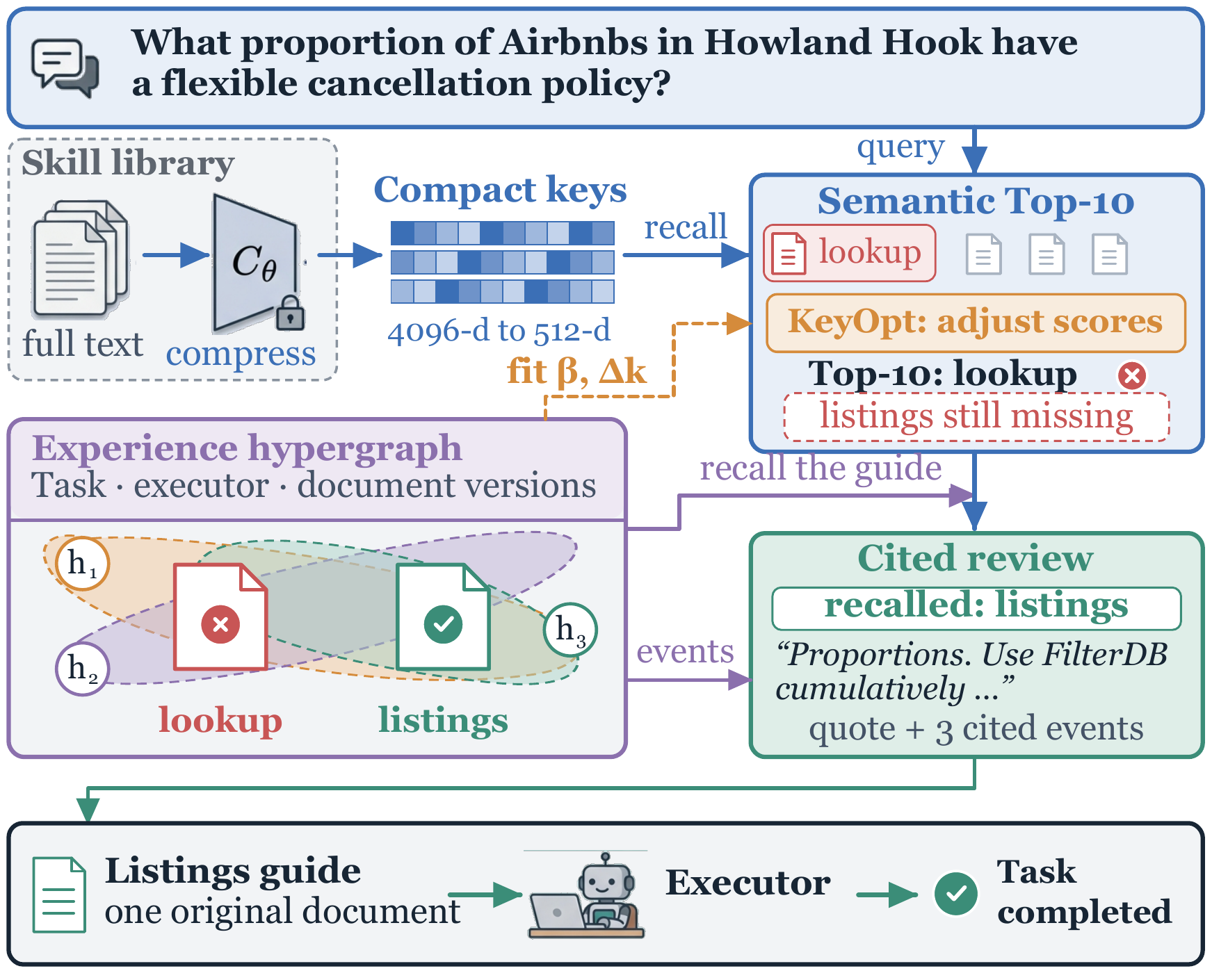}
\caption{CompSkill on ToolQA: the semantic Top-10 offers a lookup guide; events recall the missing listings guide, which KeyOpt ranks first and cited review keeps.}
\label{fig:teaser}
\end{figure}
\begin{figure*}[!t]
\centering
\includegraphics[width=\textwidth]{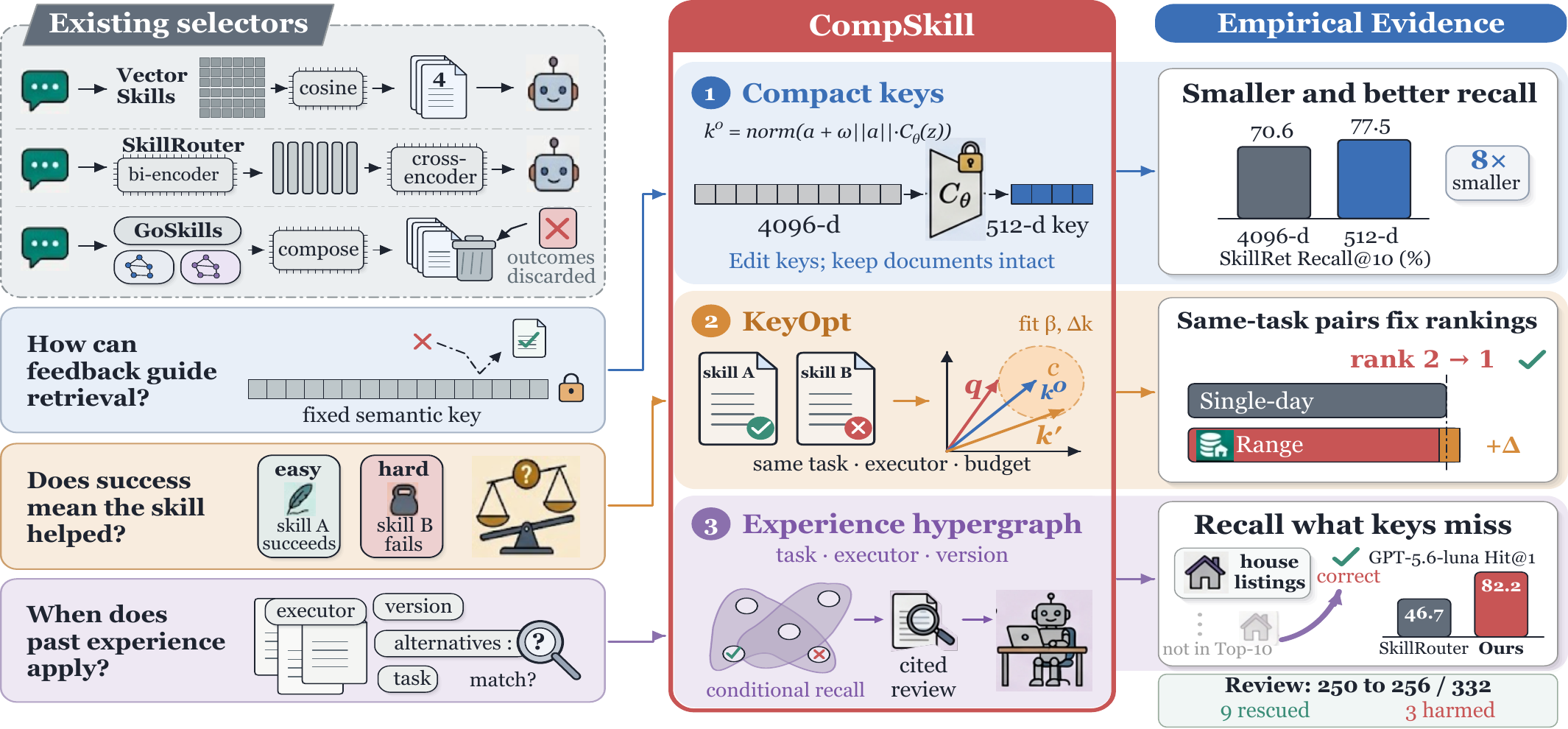}
\caption{Three challenges, CompSkill components and empirical evidence: SkillRet Recall@10 (compact keys: three-seed mean), a ToolQA Coffee ranking example, GPT-5.6-luna Hit@1 on the 332 ToolQA evaluation tasks, and paired review outcomes (9 wins, 3 losses) on the 332 newly generated tasks.}
\label{fig:baseline-overview}
\end{figure*}

Large language model (LLM) agents solve interactive tasks through tool calls \citep{yao2023react,schick2023toolformer,qin2024toolllm,li2023apibank} and reusable \emph{skills}: documents explaining how to query a database, use an API or perform a routine \citep{wang2024voyager,anthropic2025skills,agentskills2026arch,jiang2026sok}. Public skill ecosystems now hold tens of thousands of such documents \citep{li2026organizing,wang2026skillcorpus,zhang2026reusable}, and choosing a useful one has become central to execution \citep{li2026skillsbench,badhe2026survey}. Yet most retrievers choose the next skill by description alone, ignoring earlier executions.

Existing retrievers use semantic similarity, learned relevance, reranking or group structure \citep{su2026sra,kang2026skillret,zhou2025skillflow,zheng2026skillrouter,xiao2026skillsight,yang2026skillseek,wei2026beyondtopk,zeng2026goskills}. These priors organize what a skill describes; they do not establish whether a particular executor can use it under a fixed budget (Figure~\ref{fig:baseline-overview}), and a retrieved skill can even hurt execution \citep{wang2026notall,tank2026regression,jiang2026demystifying,cho2026following}. On ToolQA, even the strongest baseline supplies the labeled guide first for at most 78.3\% of tasks; a database lookup guide, for example, does not explain how to answer an aggregation request (Figure~\ref{fig:teaser}). We ask how execution feedback can teach a retriever this distinction while keeping both the skill documents and the executor unchanged.

Learning from execution raises three practical problems. \emph{Finding the relevant procedure.} Skill documents cover many details, yet small procedural differences can determine their suitability. Editing the documents to guide selection also changes the executor's instructions \citep{suzgun2025cheatsheet,zhang2025ace,ouyang2025reasoningbank,yang2026skillopt}. \emph{Comparable feedback.} Success on an easy task and failure on a hard one do not establish which skill is more useful \citep{joachims2017unbiased}; alternatives must be compared on the same task with the same executor \citep{gao2026selfevoskill}. \emph{Context for reuse.} A past outcome needs its executor, document versions and observed alternatives to inform a new choice \citep{zhao2026coalition,zhou2026eviskill}. Such experience can also recover useful skills that semantic recall missed.

\begin{figure*}[!t]
\centering
\includegraphics[width=\textwidth]{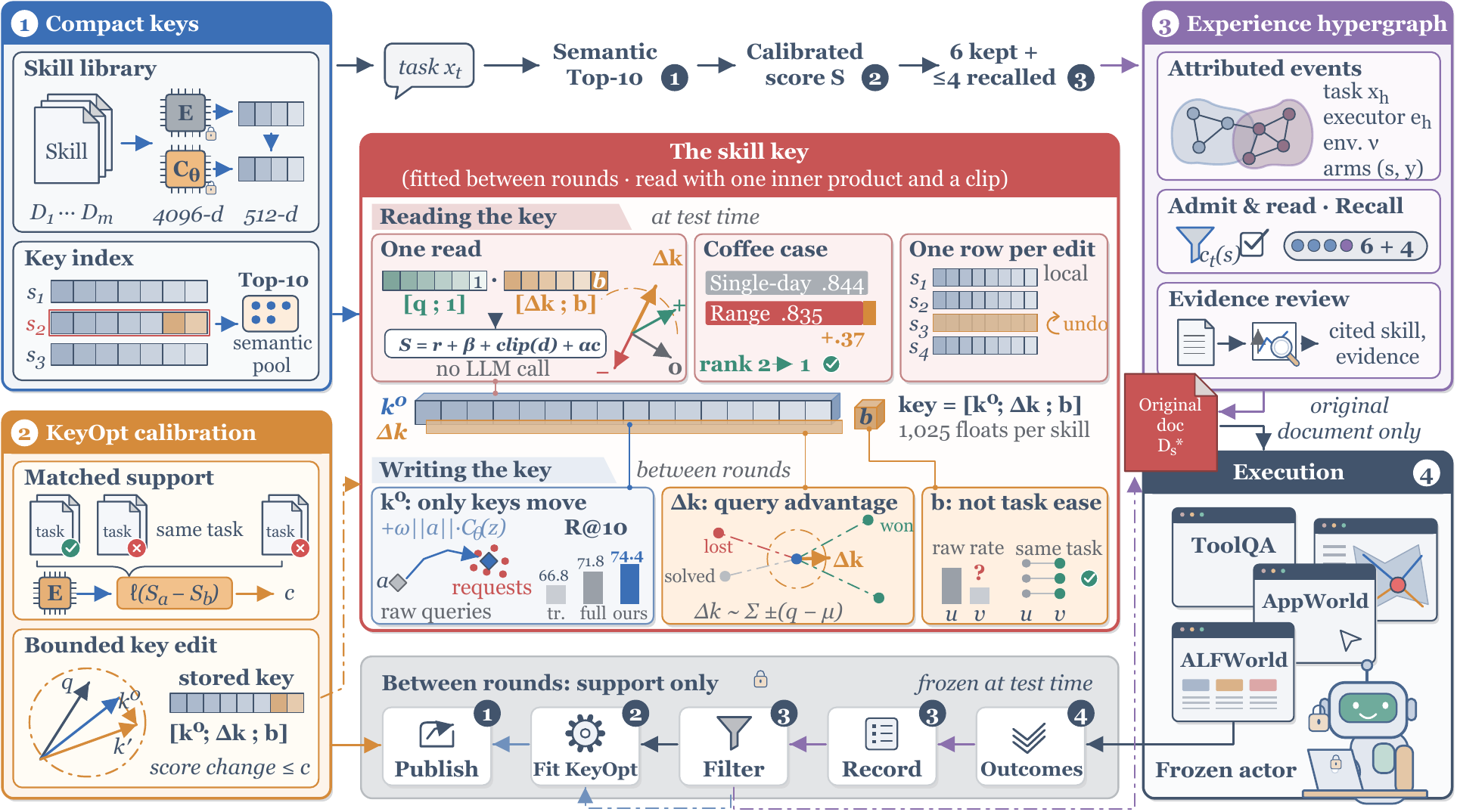}
\caption{CompSkill links compact keys, KeyOpt calibration, attributed hypergraph evidence and native execution; cited review selects an original document and the bottom loop updates only from support tasks. The center shows how a query reads one key and what experience writes into each part (Recall@10 inset: SRA library).}
\label{fig:ecr-pipeline}

\end{figure*}

To learn from execution without editing skills or the executor, we present \textbf{CompSkill}, an experience-\textbf{comp}iled \textbf{skill} retrieval framework that answers each problem with one component. \emph{Compact Semantic Indexing} gives every skill a learned 512-d key that experience can edit while the document stays intact. \emph{Matched Feedback Calibration (KeyOpt)} compiles same-task, same-executor comparisons into a key correction: a reliability offset plus a cross-validated conditional residual whose effect is capped. An \emph{Experience Hypergraph} keeps each comparison with its executor and document versions, recalls skills that semantic retrieval missed, and grounds review in cited outcomes (Figure~\ref{fig:ecr-pipeline}). Keys and events are updated only between rounds, from support runs: the executor always receives an original document, and reading the compiled experience adds one inner product per candidate and no model call to the frozen encoder and reranker.

We evaluate on SkillRet, ToolQA, AppWorld and ALFWorld \citep{kang2026skillret,zhuang2023toolqa,trivedi2024appworld,shridhar2021alfworld}. On SkillRet, 512-d compact keys outperform the 4096-d embeddings they compress (NDCG@10 70.9 versus 64.0). With one document, CompSkill completes 256/332 ToolQA tasks with GPT-5.6-luna and 260/332 with DeepSeek-V4.1-Flash, above every baseline with either executor, including BM25, SkillRet-8B, Tool2Vec and RankGPT; the GPT fits come from GPT-5.5 support and transfer to GPT-5.6-luna without refitting.
On AppWorld and ALFWorld, whose support holds few contrasting executions, it stays on par with the strongest baselines. On 180 held-out and 332 newly generated ToolQA tasks, it significantly outperforms four retrievers on each set, including Tool2Vec and RankGPT on the new tasks (Section~\ref{sec:heldout}). Error analysis shows where the gains come from: a near-miss sibling guide costs more than an unrelated document, and more than another database's guide even beside the right guide, and with the GPT fit the KeyOpt index supplies a sibling on 16 of the 180 held-out tasks where Vector Top-1 supplies 37.

\begin{table*}[!t]
\centering\small\setlength{\tabcolsep}{4pt}\renewcommand{\arraystretch}{1.06}
\begin{tabularx}{\textwidth}{@{}>{\raggedright\arraybackslash}p{2.95cm}>{\raggedright\arraybackslash}p{2.85cm}>{\raggedright\arraybackslash}X@{}}
\toprule
\textbf{Method} & \textbf{Representation} & \textbf{Selection and adaptation mechanism}\\
\midrule
Dense / Vector Skills & Document vectors & Frozen cosine similarity retrieves original documents.\\
SkillRouter / SkillRet & Encoder / reranker & Relevance pairs train recall and reranking; original skill text stays fixed.\\
NUDGE & Document vectors & Relevance labels drive constrained vector edits; the encoder stays fixed.\\
RankGPT & Candidate list & Prompted LLM comparisons reorder retrieved documents.\\
\midrule
GoS / GoSkills & Graph / skill groups & Recall prerequisites (GoS); compose role-aware groups (GoSkills).\\
SkillNet-Fabric & Typed graph + Wiki & Retrieve and expand typed relations; an Explorer selects from Wiki evidence.\\
\midrule
ExpeL / AWM / ReMe & Text memories & Runs become insights (ExpeL), workflows (AWM) or memories (ReMe).\\
SkillWeaver / SkillOpt & APIs / skill text & API synthesis (SkillWeaver); validation-gated skill-text edits (SkillOpt).\\
\midrule
\textbf{CompSkill (ours)} & \textbf{Keys + hyperedges} & \textbf{Matched outcomes} calibrate offsets and clipped 512-d key residuals with support-chosen strengths; attributed events drive recall and cited review.\\
\bottomrule
\end{tabularx}
\caption{Skill retrieval and memory methods by representation and by selection and adaptation mechanism. CompSkill combines matched key calibration with attributed evidence while retaining original skill documents.}
\label{tab:rw}
\end{table*}

\paragraph{Contributions.} (i) An architecture that compiles execution experience into retrieval while leaving documents, prompts and executor weights untouched: editable compact keys, matched calibration whose capped conditional part costs one inner product per candidate, and an experience hypergraph whose attributed events recall missed skills and ground cited review. (ii) An analysis, checked on real support data, showing that under an additive logistic outcome model matched discordant pairs identify executor-specific utility differences independently of task difficulty, that only such pairs carry information, and that the conditional correction is capped and every correction vanishes without discordant evidence (Appendix~\ref{app:theory}). (iii) Evidence from four benchmarks, two held-out ToolQA sets and a conflict test showing that gains concentrate where confusable guides decide execution and support holds contrasting outcomes.

\section{Related Work}
\label{sec:related-work}

\paragraph{Tool and skill retrieval.} Tool retrievers learn task--API relevance or traverse tool graphs \citep{qin2024toolllm,patil2024gorilla,moon2024tool2vec,qu2024colt,liu2024toolnet,shi2025toolret} on top of dense encoders \citep{karpukhin2020dpr,zhang2025qwen3embedding}. For skills, SRA and SkillRet evaluate retrieval \citep{su2026sra,kang2026skillret} and SkillRouter learns body-aware recall and reranking \citep{zheng2026skillrouter}; other work fuses fields, expands queries, contrasts neighboring skills, reasons over implicit requests or decomposes tasks \citep{goulart2026fieldaware,ding2026skill2query,wang2026neighbors,jiang2026skillreason,liu2026imagine,chen2026taskdecomp}, and routing studies probe frozen LLMs, skill composition and router-aware skill injection \citep{chen2026routerwithin,gao2026compositional,kolluru2026comparative,najjar2026surviving}. SkillNet, GoS, GoSkills and graph-based variants organize relations, dependencies and bundles \citep{liang2026skillnet,li2026gos,zeng2026goskills,xia2026grasp,li2026caskg,yao2026skilltrace,skillgraph2026,fu2026segos}. Table~\ref{tab:rw} compares these methods with CompSkill by representation and mechanism.

\paragraph{Feedback-adapted retrieval.} Pairwise ranking learns preferences from comparisons \citep{joachims2002optimizing,burges2005ranknet,joachims2017unbiased}; NUDGE, ORAG and RADEG adapt embeddings, tool vectors or utility gates \citep{zeighami2025nudge,pan2025orag,he2026radeg}. Like RADEG, KeyOpt learns from matched same-task executions, but it writes them into bounded corrections of retrieval keys rather than an execution gate, treating each key as an editable key--value entry \citep{geva2021keyvalue,meng2022rome}, and keeps the attributed events for recall and cited review. Unlike click feedback, which needs propensity corrections for presentation bias \citep{joachims2017unbiased}, matched executions of the same task cancel its difficulty under an additive logistic outcome model (Appendix~\ref{app:theory}). The compact space provides an efficient interface for execution-driven adaptation \citep{kusupati2022matryoshka}. Only retrieval parameters and evidence are adapted; the original documents and executor weights remain fixed throughout.

\paragraph{Experience and review.} Agents retain reflections, insights, workflows, procedural memories and evolving skills \citep{shinn2023reflexion,zhao2024expel,wang2024awm,cao2026reme,zheng2025skillweaver,wang2025asi,skillos2026,skill12026,he2026skillcommit,zheng2026skillprox,zhang2026memskill,wang2026skillx}, surveyed by \citet{li2026dynamicsurvey}; HyperSkill groups trajectories as hyperedges \citep{xu2026hyperskill}, and graph memories link related content \citep{edge2024graphrag,gutierrez2024hipporag}. LLMs also rerank and judge candidates \citep{sun2023rankgpt,zheng2023judging}, and skill activation can be gated by counterfactual or executability evidence \citep{guo2026skillapt,ashkenazi2026gating}. CompSkill retains comparisons with executor and document-version provenance for candidate recall and quote-and-citation review, while supplying original documents: experience changes which skill is chosen, never what the executor reads, so gains come from selection.

\section{The CompSkill Framework}

\label{sec:method}

CompSkill rests on three invariants. Experience changes only which original document is supplied, never its text or the executor. Every correction is learned from matched comparisons and traceable to the events that produced it, and its query-conditional part is capped. Reading the compiled experience costs one inner product and a clip per candidate on top of the frozen reranker score; Appendix~\ref{app:theory} formalizes and proves these properties.
\subsection{Problem Formulation}
\label{sec:task-definition}
Given a library $\mathcal S=\{s_1,\ldots,s_N\}$ of versioned documents $D_s$, a frozen executor $A_e$ and execution budget $B$, the retriever selects skills for public task context $x_t$. A snapshot $H$ of earlier support executions informs this selection:
\begin{equation}
 U_t=\pi_H(x_t;\mathcal S,e,B),\quad U_t\subseteq\mathcal S,\ |U_t|\le K.
 \label{eq:selection-task}
\end{equation}
Here $K=1$ for a single document and $K\le4$ for bundles. The executor receives the original documents in $U_t$, not the larger candidate pool from which they are selected (Section~\ref{sec:evidence}). For a trajectory $\xi$ in native environment $\mathcal E$, we maximize expected native task success
\begin{equation}
 J(\pi_H)=\mathbb E_{x\sim\mathcal D,\,\xi\sim P_{\mathcal E,A_e}^{\pi_H}(\cdot\mid x,B)}[R(\xi)],
 \label{eq:task-objective}
\end{equation}
where $P$ is the induced trajectory law, $\mathcal D$ the task distribution and $R(\xi)\in\{0,1\}$ the native success verdict. Only support runs update $H$ and the selector between rounds; both stay frozen while the evaluation tasks are executed (Figure~\ref{fig:ecr-pipeline}).

\subsection{Compact Semantic Indexing}
\label{sec:compact}

A selection key can change while its document stays intact. For each versioned document $D_s$, a frozen encoder $E$ supplies a full embedding $z_s$:
\begin{equation}
 z_s=E(D_s)\in\mathbb R^{4096}.
 \label{eq:full-embedding}
\end{equation}
A relevance-trained compressor $C_\theta$, implemented as a residual MLP, uses all coordinates to refine the 512-d anchor $a_s=z_{s,1:512}$,
\begin{equation}
 k_s^0=\mathrm{norm}\bigl(a_s+\omega\lVert a_s\rVert\,C_\theta(z_s)\bigr)\in\mathbb R^{512},
 \label{eq:compact-key}
\end{equation}
where $\mathrm{norm}(v)=v/\|v\|_2$ and $\omega$ is a learned scale. We train $C_\theta$ once on SkillRet \citep{kang2026skillret} and freeze it before feedback; target libraries receive no relevance training. Query vectors retain the normalized first 512 encoder coordinates:
\begin{equation}
 q_t=\mathrm{norm}(E(x_t)_{1:512}).
 \label{eq:query-key}
\end{equation}
Base keys retrieve the semantic pool
\begin{equation}
 \mathcal C_t^0=\operatorname{Top}_{10,s\in\mathcal S}\;q_t^\top k_s^0.
 \label{eq:base-recall}
\end{equation}

\subsection{Matched Feedback Calibration (KeyOpt)}
\label{sec:keyopt}
Compact keys rank skills by semantic relevance; KeyOpt asks with which semantically close candidate the current executor actually completes the task. It splits matched execution feedback into overall reliability, held by an offset $\beta_s$, and query-conditional suitability, held by a key residual $\Delta k_s$ whose score change is clipped:
\begin{equation}
 S(x_t,s)=r_t(s)+\beta_s+\operatorname{clip}\bigl(d_t(s),\pm c\bigr)+\alpha\,c_t(s),
 \label{eq:score}
\end{equation}
where $r_t(s)$ is the frozen SkillRouter cross-encoder score, $d_t(s)=(q_t-\mu_H)^\top\Delta k_s$ with $\mu_H$ the mean query of the support preference pairs, and $c_t(s)$ the event reading of Section~\ref{sec:evidence}. The cap $c$ and the reading weight $\alpha$ are selected on support tasks.
\paragraph{Reliability from matched wins.} On support task $i$ the executor runs two candidates under the same budget and environment; a pair $(a,b)\in P_i$ with outcomes $y_{ia}>y_{ib}$ yields a preference. For a score $f$, the ranking loss averages over the set $P$ of all such preferences,
\begin{equation}
 \mathcal L_{\mathrm{rank}}(f)=\frac{1}{|P|}\sum_{(i,a,b)\in P}\ell(f_{ia}-f_{ib}),
 \label{eq:paired-loss}
\end{equation}
with $\ell(z)=\log(1+e^{-z/\tau})$, and offsets minimize $\mathcal L_{\mathrm{rank}}(r+\beta)+\lambda\,\mathbb E_s\beta_s^2$. Under a logistic outcome model with executor-specific skill utilities $u_s$, a discordant pair is won with probability $\sigma(u_a-u_b)$ whatever the task's difficulty, so $\beta$ tracks utility differences rather than task ease (Appendix~\ref{app:theory}).
\paragraph{Conditional suitability from the same pairs.} With $\beta$ fixed, the residual is fitted to the same preferences after adding $d_i(s)$ to each score, regularized by its size and by its predictions on all support queries $H$:
\begin{equation}
 \begin{aligned}
 \mathcal L_{\Delta}={}&\mathcal L_{\mathrm{rank}}(r+\beta+d)+\lambda\,\mathbb E_s\|\Delta k_s\|_2^2\\
 &+\kappa\,\mathbb E_{i\in H,s}\bigl[d_i(s)^2\bigr].
 \end{aligned}
 \label{eq:keyopt-full}
\end{equation}
A key thus moves toward queries on which its skill beats the alternatives and away from those on which it loses; tied tasks contribute no pair. The keep term, weighted by $\kappa$, holds $d$ near zero on support queries that carry no preference, so uniform advantage stays in $\beta_s$ and the residual encodes how it varies by request (single-day versus range queries).
\paragraph{Bounded deployment with selected strengths.} Offsets and residuals are frozen before evaluation. Centering the query once per task makes $d_t(s)$ one inner product per candidate, and the clip moves no score by more than $c$ from $r_t(s)+\beta_s+\alpha c_t(s)$ (Propositions~\ref{prop:compile}--\ref{prop:bound}). Five folds of the support tasks, split by source group (tasks that share a source question), select $c$, $\kappa$, $\alpha$ and whether to use the residual, using only known support contrasts; ties prefer the weaker correction. Without discordant support pairs every correction vanishes and the index reduces to the reranker; only discordant pairs inform the corrections, and the precision of a learned utility difference grows with their number (Proposition~\ref{prop:discordance}; Figure~\ref{fig:discordance-curve}). ToolQA support keeps the residual for both executors ($c=0.3,\alpha=0$ for GPT; $c=0.2,\alpha=0.05$ for DeepSeek), whereas AppWorld and ALFWorld support, with only 0--8 discordant tasks, keeps offsets at most.

\subsection{Experience Hypergraph}
\label{sec:evidence}

\paragraph{Events as hyperedges.} We retain each comparison and its conditions as a hyperedge
\begin{equation}
 h=\bigl(x_h,e_h,\nu_h,\{(B_{hj},y_{hj})\}_j\bigr),
 \label{eq:event}
\end{equation}
binding public context $x_h$, executor $e_h$, environment $\nu_h$, and every observed arm $j$ with supplied skill set $B_{hj}$ (document versions) and outcome $y_{hj}$. Unobserved arms are omitted, and bundle outcomes stay at bundle level. A view $\mathcal H_t$ admits support events from the same executor, environment and budget whose documents are at their current versions, whose arms differ in outcome, and whose source group differs from the current task's.

\paragraph{Conditional reading.} For an admitted event $h$ that observed $s$ and at least one alternative with a different known outcome, the contrast of $s$ is its advantage over the event's mean known outcome,
\begin{equation}
 a_{h,s}=y_{h,s}-\bar y_h.
 \label{eq:event-contrast}
\end{equation}
Using the event's query embedding $q_h$, reading first selects, among the admitted events that observed $s$,
\begin{equation}
 \mathcal N_\gamma(q_t,s)=\operatorname{Top}_M\{h\in\mathcal H_t:s\in h,\ q_t^\top q_h>\gamma\},
 \label{eq:event-neighbors}
\end{equation}
where $\gamma$ is a similarity threshold and $M$ the event cap. With softmax weights $w_h$ over these similarities at temperature $T$, conditional reading gives
\begin{equation}
 c_t(s)=\sum_{h\in\mathcal N_\gamma(q_t,s)}w_h\,a_{h,s},
 \label{eq:reading}
\end{equation}
and an empty neighborhood gives zero.
\paragraph{Event recall.} The candidate pool keeps the top six semantic candidates and adds up to four skills from outside the semantic Top-10 that won in similar admitted events, so skills that succeeded on similar tasks can enter even when their description is a weak match. Candidates are ranked by Equation~\ref{eq:score}; in bundle settings GoSkills' composer forms bundles from this pool, and bundle hyperedges add corrections at the level of whole bundles.
\paragraph{Cited review.} A frozen reviewer from the executor's model family receives the task, the top eight candidate documents (four bundles), the calibrated proposal and at most six hyperedges from the candidates' neighborhoods $\mathcal N_\gamma(q_t,s)$. It must return a candidate, an exact quote from its document and the IDs of any cited events (Appendix~\ref{app:prompts}); invalid outputs fall back to the proposal. Only the selected document reaches the executor. Reading hyperparameters ($\gamma=0.3$, $M=16$, $T=0.1$) are fixed across benchmarks and executors; $\alpha$ is chosen by cross-validation with the other strengths.

\paragraph{Rounds.} The admission rule also selects the events that fit KeyOpt; on ToolQA with GPT-5.5, 72 of the 318 support tasks pass it. Algorithm~\ref{alg:compskill} (Appendix~\ref{app:keyopt-details}) gives one round: select and execute, then record, filter, fit and publish a snapshot for the next round. We call the reranker with fitted corrections the KeyOpt index; on ToolQA its pool holds the semantic Top-10 and up to four recalled skills, while CompSkill keeps six semantic and four recalled candidates and adds cited review.

\begin{table*}[!t]
\centering\small\setlength{\tabcolsep}{3.5pt}\renewcommand{\arraystretch}{1.06}
\begin{tabular*}{\textwidth}{@{\extracolsep{\fill}}cl c cccc ccc@{}}
\toprule
 & & & \multicolumn{4}{c}{Success} & \multicolumn{3}{c}{CompSkill versus baseline}\\
\cmidrule(lr){4-7}\cmidrule(l){8-10}
Executor & Method & \#Docs & Tasks & Rate (\%) & Complex (\%) & Worst DB (\%) & Gain [95\% CI] & W/L & $p$\\
\midrule
\multirow{10}{*}{\rotatebox[origin=c]{90}{\textbf{DeepSeek-V4.1-Flash}}} & No Skills & 0 & 200 & 60.2 & 37.1 & 41.4 & $+$60 [$42$, $77$] & 75/15 & $<$0.001\\
\cmidrule(l){2-10}
 & Vector Top-4 & 4 & 240 & 72.3 & 59.0 & 51.4 & $+$20 [$7$, $34$] & 34/14 & 0.006\\
 & GoSkills & 2.57 & 218 & 65.7 & 43.8 & 53.3 & $+$42 [$26$, $59$] & 59/17 & $<$0.001\\
\cmidrule(l){2-10}
 & Vector Top-1 & 1 & 236 & 71.1 & 57.3 & 55.7 & $+$24 [$13$, $36$] & 30/6 & $<$0.001\\
 & SkillRouter & 1 & 226 & 68.1 & 52.8 & 51.4 & $+$34 [$21$, $47$] & 41/7 & $<$0.001\\
 & BM25 & 1 & 219 & 66.0 & 45.5 & 54.3 & $+$41 [$27$, $55$] & 50/9 & $<$0.001\\
 & SkillRet-8B & 1 & 233 & 70.2 & 55.6 & 55.7 & $+$27 [$15$, $40$] & 34/7 & $<$0.001\\
 & Tool2Vec & 1 & 252 & 75.9 & 65.2 & 58.6 & $+$8 [$-1$, $17$] & 14/6 & 0.115\\
 & RankGPT & 1 & 243 & 73.2 & 60.1 & 55.7 & $+$17 [$8$, $26$] & 20/3 & $<$0.001\\
\cmidrule(l){2-10}
 & \textbf{CompSkill (ours)} & 1 & \textbf{260} & \textbf{78.3} & \textbf{67.4} & \textbf{65.7} & --- & --- & ---\\
\midrule
\multirow{10}{*}{\rotatebox[origin=c]{90}{\textbf{GPT-5.6-luna}}} & No Skills & 0 & 197 & 59.3 & 36.5 & 41.4 & $+$59 [$41$, $77$] & 75/16 & $<$0.001\\
\cmidrule(l){2-10}
 & Vector Top-4 & 4 & 245 & 73.8 & 61.8 & 58.6 & $+$11 [$-2$, $23$] & 26/15 & 0.117\\
 & GoSkills & 2.57 & 221 & 66.6 & 50.0 & 52.9 & $+$35 [$20$, $50$] & 49/14 & $<$0.001\\
\cmidrule(l){2-10}
 & Vector Top-1 & 1 & 235 & 70.8 & 58.4 & 55.7 & $+$21 [$10$, $33$] & 28/7 & $<$0.001\\
 & SkillRouter & 1 & 225 & 67.8 & 54.5 & 54.3 & $+$31 [$18$, $44$] & 39/8 & $<$0.001\\
 & BM25 & 1 & 223 & 67.2 & 49.4 & 57.1 & $+$33 [$19$, $47$] & 44/11 & $<$0.001\\
 & SkillRet-8B & 1 & 226 & 68.1 & 50.6 & 44.3 & $+$30 [$18$, $42$] & 36/6 & $<$0.001\\
 & Tool2Vec & 1 & 238 & 71.7 & 55.6 & 50.0 & $+$18 [$7$, $30$] & 27/9 & 0.004\\
 & RankGPT & 1 & 249 & 75.0 & 61.2 & 55.7 & $+$7 [$-1$, $15$] & 12/5 & 0.143\\
\cmidrule(l){2-10}
 & \textbf{CompSkill (ours)} & 1 & \textbf{256} & \textbf{77.1} & \textbf{65.7} & \textbf{61.4} & --- & --- & ---\\
\bottomrule
\end{tabular*}
\caption{ToolQA results ($n=332$). Rows group No Skills, multi-document and single-document baselines; Docs: documents supplied per task (mean for GoSkills). Gain: additional tasks CompSkill completes, with a paired task-bootstrap 95\% interval; W/L: tasks that only CompSkill or only the baseline completes; $p$: two-sided exact McNemar test. Complex: 178 complex queries; Worst DB: the weakest of four databases. Best in bold. Breakdown by query type, database and first-choice accuracy (Hit@1): Appendix~\ref{app:more-results}.}
\label{tab:main}

\end{table*}
\section{Experiments}
\label{sec:experiments}
\subsection{Setup}
\paragraph{Skill retrieval.} The SkillRet test split contains 6,006 skills and 4,392 queries \citep{kang2026skillret}. Only the compressor learns from TRAIN; encoders stay frozen. We compare native, truncated and compact vectors, plus lexical and dense references (Table~\ref{tab:skillret}). NDCG, Recall and Completeness@10 measure relevance without execution; seeds, skill-tuned references and controls are in Appendix~\ref{app:skillret}.
\paragraph{Execution benchmarks.} ToolQA has 332 tasks and 785 documents (eight database guides and 777 public skills), and uses the released ReAct agent with native tools, 20 actions and 900\,s \citep{zhuang2023toolqa}. AppWorld uses 90 Test-Challenge tasks from 30 templates (selected by template-ID hash) and a library of 469 API documents; support and test apps are disjoint \citep{trivedi2024appworld}. ALFWorld uses all 134 \texttt{valid\_unseen} games and 37 skills \citep{shridhar2021alfworld}. Both allow 30 actions and 900\,s. Native evaluators determine task success on each of the three benchmarks.
\paragraph{Executors and feedback.} DeepSeek-V4.1-Flash and the GPT executors use low reasoning. Each executor's KeyOpt is fitted on its own disjoint support tasks (ToolQA $318\times2$ documents; AppWorld $90\times2$; ALFWorld 12 games, each with two skills and two bundles) and frozen before evaluation. ToolQA is executed by GPT-5.6-luna with the fits and events of GPT-5.5 support, so every GPT result on ToolQA also tests transfer to a newer executor; AppWorld and ALFWorld use GPT-5.5. DeepSeek fits come from September support, and its results combine executions from checkpoints served from September to late October (Appendix~\ref{app:post}). Section~\ref{sec:rq6} tests transfer across executor versions.
\paragraph{Execution comparisons.} We compare No Skills; Vector Skills, which supplies the one or four nearest complete documents by frozen 4096-d cosine (Vector Top-1, Top-4); official GoSkills group retrieval with full-document rendering \citep{zeng2026goskills}; and full-library SkillRouter recall and reranking \citep{zheng2026skillrouter}. On ToolQA we add BM25, SkillRet-8B \citep{kang2026skillret}, Tool2Vec, which averages the queries a skill solved in support \citep{moon2024tool2vec}, and RankGPT listwise reranking \citep{sun2023rankgpt}. Environments, tools and action budgets are matched within each executor; baselines keep their released settings (Appendix~\ref{app:baseline-contracts}).
\paragraph{Metrics.} We report task success, first-choice accuracy (Hit@1: the first supplied document is the labeled guide), AppWorld scenario goal completion (SGC: all three tasks of a template succeed), and ToolQA success on 178 complex queries and on its weakest domain. Exact McNemar tests compare paired outcomes on the same tasks.
\subsection{RQ1: Main Results}
\label{sec:rq1}\label{sec:main-results}
\begin{figure*}[!t]
\centering
\includegraphics[width=\textwidth]{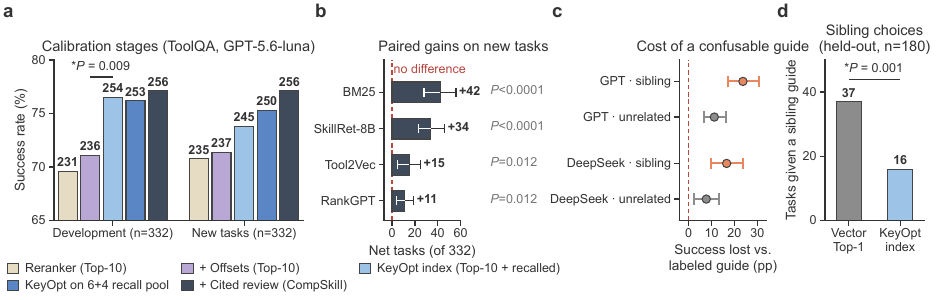}
\caption{(a) ToolQA successes as calibration stages are added (GPT-5.6-luna; Development: the 332 evaluation tasks); from the KeyOpt index on, the pool also holds skills recalled by the event store; bracket: prespecified Holm-corrected test. (b) Net gains of CompSkill over the four prespecified baselines on the 332 new tasks, with 95\% bootstrap intervals and Holm-corrected exact McNemar $P$. (c) Within-task success lost on the 332 evaluation tasks when a sibling or an unrelated document replaces the labeled guide (GPT: GPT-5.5). (d) Sibling guides supplied on the 180 held-out tasks by Vector Top-1 and the KeyOpt index (GPT fit).}
\label{fig:component-evidence}
\vspace{3pt}
\includegraphics[width=\textwidth]{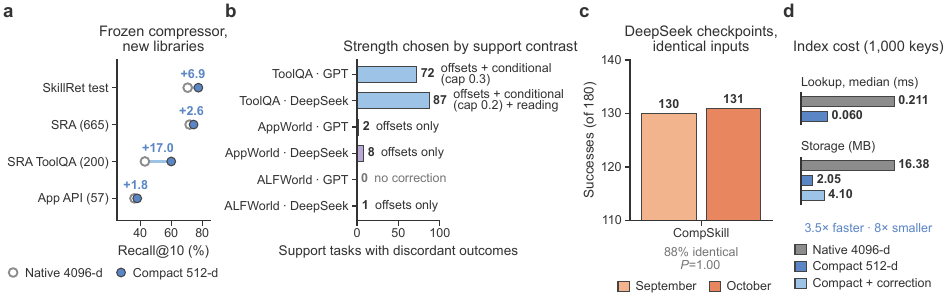}
\caption{(a) Recall@10 of the frozen compressor on libraries it was not trained on. (b) Support tasks with discordant outcomes and the correction selected by cross-validation. (c) CompSkill under two DeepSeek checkpoints on identical inputs (180 held-out tasks). (d) Cached lookup latency and key storage for 1,000 keys.}
\label{fig:behavior-evidence}
\end{figure*}

\paragraph{Retrieval.} Compact keys improve SkillRet NDCG@10 from 64.0 to 70.9 and Recall@10 from 70.6\% to 77.5\%, using 512 rather than 4,096 dimensions of the same encoder and text (Figure~\ref{fig:behavior-evidence}a).

\setcounter{dbltopnumber}{1}
\paragraph{Execution.} On ToolQA, CompSkill supplies one document and completes 256/332 tasks with GPT-5.6-luna and 260/332 with DeepSeek-V4.1-Flash (Table~\ref{tab:main}), more than every baseline with either executor, including four-document Vector Top-4, while supplying about a fifth of its guidance tokens (1.1k versus 5.1k with GPT-5.6-luna; Appendix~\ref{app:heldout}). It also leads on complex queries (65.7\% and 67.4\%) and on the weakest domain (61.4\% and 65.7\%). With GPT-5.6-luna it exceeds BM25, SkillRet-8B and Tool2Vec (44/11, 36/6 and 27/9 wins/losses, Holm-corrected $p\le0.008$), gains 21 tasks over Vector Top-1 ($p<0.001$) and leads RankGPT (249). With DeepSeek it exceeds BM25, SkillRet-8B and RankGPT (50/9, 34/7 and 20/3, Holm-corrected $p\le0.001$), gains 24 tasks over Vector Top-1 ($p<0.001$) and leads Tool2Vec (252). With GPT-5.6-luna it also supplies the labeled guide first most often (82.2\% versus 78.0\% for RankGPT, the next best; Appendix~\ref{app:more-results}); with DeepSeek it departs from the labeled guide more often than RankGPT (74.7\% versus 78.3\%) yet completes 17 more tasks (20/3 wins/losses), because its calibration follows what this executor completes.

\begin{table}[t]
\centering\footnotesize\setlength{\tabcolsep}{3pt}\renewcommand{\arraystretch}{1.05}
\begin{tabular}{@{}lccccccc@{}}
\toprule
 & \multicolumn{2}{c}{ToolQA} & \multicolumn{2}{c}{AppWorld} & \multicolumn{2}{c}{ALFWorld} & \\
\cmidrule(lr){2-3}\cmidrule(lr){4-5}\cmidrule(lr){6-7}
Method & DS & GPT & DS & GPT & DS & GPT & Total\\
\midrule
\textit{Support contrasts} & \textit{87} & \textit{72} & \textit{8} & \textit{2} & \textit{1} & \textit{0} & \\
\midrule
No Skills & 200 & 197 & 68 & 78 & 118 & 125 & 786\\
Vector Top-4 & 240 & 245 & 70 & 78 & 123 & 130 & 886\\
Vector Top-1 & 236 & 235 & 69 & \textbf{81} & \textbf{127} & 128 & 876\\
GoSkills & 218 & 221 & \textbf{74} & \textbf{81} & 124 & 130 & 848\\
SkillRouter & 226 & 225 & 65 & 80 & 126 & 130 & 852\\
\midrule
\rowcolor{tint}\textbf{CompSkill} & \textbf{260} & \textbf{256} & \textbf{74} & \textbf{81} & 124 & \textbf{131} & \textbf{926}\\
\bottomrule
\end{tabular}
\caption{Successes of the methods run on all three execution benchmarks (DS: DeepSeek-V4.1-Flash; GPT: GPT-5.6-luna with the GPT-5.5 fits on ToolQA, GPT-5.5 elsewhere). ToolQA: 332 tasks (Table~\ref{tab:main}); AppWorld: task completion on 90 Challenge tasks; ALFWorld: 134 \texttt{valid\_unseen} games; Total: 1,112 tasks. Support contrasts: support tasks whose two executed documents differ in outcome; with eight or fewer, cross-validation keeps offsets at most, and AppWorld uses the calibrated index without recall or review. AppWorld SGC (of 30 templates), by row: DS 14, 14, 18, 20, 16, 19; GPT 22, 23, 25, 25, 23, 25. Bold: best per column.}
\label{tab:transfer}
\end{table}

Table~\ref{tab:transfer} places ToolQA beside AppWorld and ALFWorld for the methods run on all three, together with the number of support tasks whose two executed documents differ in outcome. CompSkill leads most where support holds the most such tasks, as expected when only discordant pairs inform calibration (Proposition~\ref{prop:discordance}, Figure~\ref{fig:discordance-curve}). ToolQA support holds 87 (DeepSeek) and 72 (GPT), and CompSkill leads every shared baseline by at least 20 and 11 tasks, respectively; AppWorld and ALFWorld support hold eight or fewer, and CompSkill stays on par with the strongest baselines. Over all 1,112 tasks it completes 926, 40 more than the next method, four-document Vector Top-4 (886).

AppWorld support holds two and eight discordant tasks, so cross-validation keeps offsets only, and because Challenge apps never occur in support, these change none of the 90 Challenge selections. The calibrated index matches the strongest baselines with both executors (81/90 with GPT-5.5, 74/90 with DeepSeek), whereas SkillRouter with DeepSeek falls below No Skills (65 versus 68). On ALFWorld, CompSkill reaches 131/134 with GPT-5.5, the best result. With DeepSeek, whose twelve support games yield a single discordant comparison, it completes 124, level with GoSkills and within three games of every skill-supplying method (123--127), while No Skills already completes 118. The benchmark leaves little room for selection: all six methods solve 104 of the 134 games and none solves 3, so they differ on only 27. KeyOpt compiles only the contrasts that support records (Section~\ref{sec:keyopt}): on ToolQA, offsets and the KeyOpt index add 23 and 10 successes on the 332 evaluation and 332 new tasks with GPT-5.6-luna (Section~\ref{sec:rq2}), whereas here they have almost nothing to compile. Over the four AppWorld and ALFWorld settings, CompSkill completes the most tasks (410 of 448), ahead of GoSkills (409) and Vector Top-1 (405).

\subsection{Held-out and New Tasks}
\label{sec:heldout}
\begin{table}[t]
\centering\small\setlength{\tabcolsep}{4pt}
\begin{tabular}{@{}lcrrr@{}}
\toprule
Method & Docs & Eval. & New & All\\
 & & $n{=}332$ & $n{=}332$ & $n{=}844$\\
\midrule
Vector Top-1 & 1 & 235 & 227 & 568\\
SkillRouter & 1 & 225 & 225 & 556\\
BM25 & 1 & 223 & 214 & 543\\
SkillRet-8B & 1 & 226 & 222 & 553\\
RankGPT & 1 & 249 & 245 & 612\\
Tool2Vec & 1 & 238 & 241 & 604\\
Vector Top-4 & 4 & 245 & 234 & 605\\
\midrule
KeyOpt index & 1 & 254 & 245 & 618\\
\rowcolor{tint}\textbf{CompSkill} & 1 & \textbf{256} & \textbf{256} & \textbf{636}\\
\bottomrule
\end{tabular}
\caption{ToolQA successes with GPT-5.6-luna and the GPT-5.5 fits on the 332 evaluation tasks (Table~\ref{tab:main}) and 332 newly generated tasks; All adds the 180 held-out tasks of Appendix~\ref{app:heldout}. Bold: best in each column.}
\label{tab:heldout}
\end{table}

We further evaluate, with every index, event store, reviewer prompt and baseline frozen in advance, on 332 new tasks from the official ToolQA templates and the remaining 180 SRA ToolQA tasks for the same eight guides (Appendix~\ref{app:heldout}).

On the 332 new tasks CompSkill completes 256, more than every baseline, and significantly exceeds RankGPT (245), Tool2Vec (241), SkillRet-8B (222) and BM25 (214), with 13/2 to 49/7 wins/losses (Holm-corrected $p\le0.012$; Figure~\ref{fig:component-evidence}b). On the 180 held-out tasks it significantly exceeds Vector Top-1, SkillRouter, BM25 and SkillRet-8B ($p\le0.006$; Appendix~\ref{app:heldout}). Over all 844 ToolQA tasks executed with GPT-5.6-luna, CompSkill completes 636, 24 more than the strongest baseline (RankGPT, 612) and 31 more than four-document Vector Top-4 (Table~\ref{tab:heldout}).

\subsection{RQ2: Ablation Study}
\label{sec:rq2}
Figure~\ref{fig:component-evidence}a adds the calibration stages one at a time with GPT-5.6-luna, starting from the reranker over the compact-key Top-10 (keys are compared on retrieval in Section~\ref{sec:rq3}). Offsets add 5 and 2 successes on the 332 evaluation and 332 new tasks, and the KeyOpt index 18 and 8 more, significantly on the evaluation tasks (26/8 wins/losses, Holm-corrected $p=0.009$). This step adds the clipped conditional term together with the recalled candidates that the index also scores. Keeping six semantic and four recalled candidates and adding cited review brings CompSkill to 256 on both task sets.

\definecolor{csnavy}{HTML}{1D3657}\definecolor{cspurple}{HTML}{714965}\definecolor{csred}{HTML}{7D261E}\definecolor{csgold}{HTML}{AE8335}\definecolor{csgreen}{HTML}{4B8660}\definecolor{csrust}{HTML}{A55039}\definecolor{csink}{HTML}{202834}
\begin{table*}[!t]
\centering
\begin{tikzpicture}[x=1pt,y=1pt,font=\footnotesize,
  pnl/.style={draw=#1,fill=#1!7,rounded corners=3pt,line width=.9pt},
  ttl/.style={font=\footnotesize\bfseries,text=white,anchor=west,inner sep=0pt},
  badge/.style={circle,fill=white,text=#1,font=\footnotesize\bfseries,inner sep=1pt,minimum size=13pt},
  chip/.style={draw=csnavy!45,line width=.5pt,fill=white,rounded corners=1.5pt,
    minimum height=11pt,minimum width=70pt,text width=68pt,align=left,inner sep=1pt,anchor=west},
  ev/.style={draw=cspurple,fill=cspurple!16,line width=.6pt,rounded corners=3pt,
    inner sep=1pt,align=center,minimum width=70pt},
  guide/.style={rounded corners=2pt,inner sep=1pt,align=center,minimum width=52pt},
  flow/.style={-{Stealth[length=4pt]},line width=1.1pt,draw=csink!85}]
\useasboundingbox (0,0) rectangle (453.54,118.87);
\fill[csnavy!12,rounded corners=2pt] (0,105) rectangle (453.54,118.87);
\fill[csnavy,rounded corners=1pt] (0,105) rectangle (2,118.87);
\node[anchor=west,inner sep=0pt] at (5,111.94) {\includegraphics[width=.38cm]{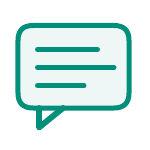}};
\node[anchor=west,inner sep=0pt] at (18,111.94) {\emph{What proportion of Airbnbs in Howland Hook have a flexible cancellation policy?}};
\draw[pnl=csnavy] (0,0) rectangle (150.24,100.77);
\fill[csnavy,rounded corners=3pt] (0,85.77) rectangle (150.24,100.77);
\fill[csnavy] (0,85.77) rectangle (150.24,94);
\node[badge=csnavy] at (8,93.27) {1};
\node[ttl] at (18,93.27) {Top-10 + KeyOpt};
\foreach \y/\r/\n in {78/1/Vrbo,66/2/Airbnb (web),54/3/Skiplagged,42/4/Hotels.com,30/5/Booking.com}{
  \node[chip] at (3,\y) {{\color{muted}\r}\ \n};}
\foreach \y/\r/\n in {66/7/Skiplagged,54/8/Expedia,42/9/hotel (zh),30/10/Hertz}{
  \node[chip] at (77,\y) {{\color{muted}\r}\ \n};}
\node[chip,fill=csrust!12,draw=csrust!80] at (77,78) {{\color{muted}6}\ \textbf{Lookup} \textcolor{csrust}{$\times$}};
\node[anchor=west,inner sep=0pt,text=csink] at (3,17) {Reranker: lookup $\times$; listings absent};
\node[anchor=west,inner sep=0pt,text=muted] at (3,5) {KeyOpt: listings $.60$, lookup $-.16$};
\draw[pnl=cspurple] (157.32,0) rectangle (296.22,100.77);
\fill[cspurple,rounded corners=3pt] (157.32,85.77) rectangle (296.22,100.77);
\fill[cspurple] (157.32,85.77) rectangle (296.22,94);
\node[badge=cspurple] at (165.32,93.27) {2};
\node[ttl] at (175.32,93.27) {Cited hyperedges};
\node[anchor=west,inner sep=0pt,text=muted] at (161,79) {3 matched support events};
\foreach \y in {62,37,12}{
  \draw[csrust!75,line width=.6pt,dashed] (232,\y) -- (243,61);
  \draw[csgreen,line width=.75pt] (232,\y) -- (243,24);}
\node[ev] (e1) at (196,62) {Fordham};
\node[ev] (e2) at (196,37) {Stuyvesant T.};
\node[ev] (e3) at (196,12) {West Brighton};
\node[guide,draw=csgold!70,fill=csgold!14] (lk) at (269,61) {DB lookup\\\textcolor{csrust}{$\times$ 3/3}};
\node[guide,draw=csgreen!70,fill=csgreen!12] (ls) at (269,24) {DB listings\\\textcolor{csgreen}{\checkmark\ 3/3}};
\draw[pnl=csred] (303.31,0) rectangle (453.54,100.77);
\fill[csred,rounded corners=3pt] (303.31,85.77) rectangle (453.54,100.77);
\fill[csred] (303.31,85.77) rectangle (453.54,94);
\node[badge=csred] at (311.31,93.27) {3};
\node[ttl] at (321.31,93.27) {Recall + cited review};
\node[anchor=west,inner sep=0pt] at (307,79) {Recalls listings + flights};
\draw[fill=white,draw=black!30,rounded corners=1.5pt] (307,23) rectangle (449,72);
\node[anchor=west,font=\footnotesize\bfseries,inner sep=0pt] at (310,65) {Airbnb listings guide};
\node[anchor=west,text=muted,inner sep=0pt] at (310,53) {Derived calculations};
\node[anchor=north west,fill=csgreen!18,inner sep=1pt,text width=134pt,align=left] at (309,47) {``\textbf{Proportions.} Use FilterDB \mbox{cumulatively} \ldots''};
\node[anchor=west,inner sep=0pt,text=muted] at (307,17) {Review cites all 3 events};
\node[anchor=west,inner sep=0pt] at (307,5) {{\color{csgreen}\checkmark} GPT-5.6-luna\quad {\color{csgreen}\checkmark} DeepSeek};
\draw[flow] (150.7,50) -- (156.7,50);
\draw[flow] (296.7,50) -- (302.7,50);
\end{tikzpicture}\par\vspace{3pt}
\let\ok\yes
\footnotesize\setlength{\tabcolsep}{4pt}\renewcommand{\arraystretch}{1.0}
\begin{tabularx}{\textwidth}{@{}p{3.35cm}X p{3.9cm}>{\centering\arraybackslash}p{.62cm}>{\centering\arraybackslash}p{.62cm}@{}}
\toprule
Selector & Supplied guidance & Signal behind the choice & Luna & DS\\
\midrule
SkillRouter & a web hotel-booking skill & learned relevance & \no & \no\\
BM25 & a Skiplagged travel skill & lexical overlap & \no & \no\\
Vector Top-1 & DB lookup guide & embedding similarity & \no & \no\\
SkillRet-8B, RankGPT & DB lookup guide & dense retrieval, LLM reranking & \no & \no\\
Tool2Vec & DB listings guide & queries solved in support & \ok & \ok\\
Vector Top-4 & lookup, Vrbo, listings, web skill (4 docs) & similarity, four documents & \ok & \ok\\
GoSkills & listings, lookup, a web skill (3 docs) & skill-group structure & \ok & \ok\\
\midrule
Reranker + offsets & DB lookup guide & Top-10 only & \no & \no\\
KeyOpt index & DB listings guide (recalled) & offsets + conditional term & \ok & \ok\\
\rowcolor{tint}\textbf{CompSkill} & \textbf{DB listings guide} & recall + review citing 3 events & \ok & \ok\\
\bottomrule
\end{tabularx}
\caption{Reference ToolQA task 00348. Three matched support events, in which the lookup guide failed and the listings guide succeeded, recall the listings guide that the Top-10 misses; KeyOpt ranks it first (reranker, offset and clipped conditional term: $.335{-}.032{+}.300$ versus $.107{+}.032{-}.300$ for the lookup guide, GPT fit), cited review keeps it, and both executors succeed. Quote and citations are from the DeepSeek review; Luna: GPT-5.6-luna.}
\label{tab:case}\label{fig:case}
\end{table*}

\subsection{RQ3: Why Do Compact Keys Help?}
\label{sec:rq3}
The compressor learns a retrieval representation from all 4,096 input coordinates; it does not shorten the skill document. At 512 dimensions, truncation and random projection reach 62.2 and 60.8 NDCG, versus 70.9 for learned keys. The gain thus comes from learned relevance, not fewer dimensions. Coverage improves beyond the first relevant match: Completeness@10 rises from 28.7\% to 39.9\% on 1,838 multi-skill queries with no relevant TRAIN name/text match (Appendix~\ref{app:skillret}).

\subsection{RQ4: How Does Feedback Rerank?}
\label{sec:rq4}
Calibration acts on the errors that descriptions resolve least reliably. Within the same ToolQA task, supplying a sibling guide for the right database instead of the labeled guide costs 17--25 points of success, more than an unrelated document (8--16 points; Figure~\ref{fig:component-evidence}c and Table~\ref{tab:anatomy}). Even beside the labeled guide, a sibling costs more than another database's guide: on 96 held-out tasks, GPT-5.6-luna completes 67 with the labeled guide and its sibling and 75 with another database's guide (exact $p=0.039$; Appendix~\ref{app:anatomy}), the kind of partial conflict that induces procedural hallucinations \citep{jiang2026conflict}. Both guides describe the same database, so description similarity separates them poorly, whereas matched support comparisons record which one this executor completes.

With the GPT fit, the KeyOpt index supplies a sibling on 16 of the 180 held-out tasks against 37 for Vector Top-1: it removes the sibling on 31 tasks and introduces one on 10 (exact $p=0.001$; Figure~\ref{fig:component-evidence}d). On a Coffee range query, the reranker prefers the single-day guide (0.844 versus 0.835); the GPT fit lowers it by 0.37 and raises the range guide by 0.37, 0.26 of each from the conditional term, and the executor fails with the former but succeeds with the latter (Figures~\ref{fig:baseline-overview} and~\ref{fig:ecr-pipeline}).

Cross-validation also decides how much correction the support can justify (Figure~\ref{fig:behavior-evidence}b). ToolQA support holds 72 (GPT) and 87 (DeepSeek) tasks with discordant outcomes, and both fits keep a clipped conditional term. AppWorld and ALFWorld support hold at most eight such tasks; there cross-validation keeps offsets at most, and where support shows no contrast at all the index reduces to the reranker. On support data, matched predictions improve steadily with the number of discordant tasks, whereas concordant outcomes alone point the wrong way (Appendix~\ref{app:theory}, Figure~\ref{fig:discordance-curve}).

\subsection{RQ5: How Does Experience Help?}
\label{sec:rq5}
Events contribute in two places: recalled skills enter the KeyOpt index pool together with the conditional term (Section~\ref{sec:rq2}), and cited review lets the executor's model family check the calibrated proposal against similar past comparisons. On the new tasks, keeping six semantic and four recalled candidates adds 5 successes over the KeyOpt index and review 6 more (9/4 and 9/3 wins/losses), raising the total from 245 to 256; on the evaluation tasks, where the KeyOpt index already completes 254, they add two more. Review must justify any switch with an exact quote from the chosen document.

\subsection{RQ6: What Transfers Across Settings?}
\label{sec:rq6}
Without target-library fitting, the frozen compressor improves Recall@10 on SRA, on its 200 ToolQA queries (43.0\% to 60.0\%) and on 57 cached AppWorld API requests (Figure~\ref{fig:behavior-evidence}a).

Calibration also transfers across executor versions without refitting. The ToolQA fits come from GPT-5.5 support, yet with GPT-5.6-luna CompSkill leads every baseline on the new tasks (Section~\ref{sec:heldout}). With a DeepSeek checkpoint released after the fits, CompSkill completes 131 of the 180 held-out tasks (130 with the original checkpoint), and its identical inputs give the same outcome under both checkpoints on 88\% of tasks, with no systematic shift in either direction (Figure~\ref{fig:behavior-evidence}c; Appendix~\ref{app:heldout}).

\subsection{RQ7: What Does Selection Cost?}
\label{sec:rq7}
Compact keys make the semantic lookup cheap: with 1,000 cached keys, median Top-10 lookup falls from 0.211 to 0.060\,ms (3.5$\times$), and base-key storage from 16.38 to 2.05\,MB (8$\times$; Figure~\ref{fig:behavior-evidence}d). Compiled experience adds one inner product and a clip per candidate and no model call; beyond the query encoder and the frozen reranker, only review adds a model call, once per task.

\subsection{RQ8: Case Study}
\label{sec:rq8}
Table~\ref{tab:case} follows a request for the proportion of Airbnb listings in Howland Hook with a flexible cancellation policy. Vector Top-1, SkillRet-8B, RankGPT and the reranker with offsets supply a guide for single-record lookups, SkillRouter and BM25 generic web skills, and both executors fail. The listings guide, which covers aggregation over listings, is absent from the semantic Top-10; the event store recalls it because similar support tasks succeeded with it and failed with the lookup guide. In the pool it already outranks the lookup guide by reranker score (0.335 versus 0.107), and KeyOpt's clipped conditional term, learned from those comparisons, widens the margin ($+0.30$ versus $-0.30$ with the GPT fit). The KeyOpt index and CompSkill supply only this document, and both executors succeed. Tool2Vec, which also scores skills by the support queries they solved, finds the same guide; the other baselines that succeed supply three or four documents. One well-chosen guide does what three or four documents do, without bringing in conflicting procedures (Appendix~\ref{app:anatomy}).

\section{Conclusion}
CompSkill compiles matched execution comparisons into reliability offsets and capped residuals stored with each compact skill key, and keeps the comparisons as attributed events that recall missed skills and ground cited review, while documents and executors stay fixed. Experience therefore changes which original document the executor receives, never its text, and reading it costs one inner product per candidate. With one document CompSkill leads every baseline on the 332 ToolQA evaluation tasks with both executors, significantly outperforms BM25, SkillRet-8B, Tool2Vec and RankGPT on newly generated tasks, and over all 844 ToolQA tasks run with GPT-5.6-luna completes 24 more than the strongest baseline. Its gains concentrate where confusable guides decide whether execution succeeds and support records enough contrasting executions to separate them; on AppWorld and ALFWorld, where support records few, it stays on par with the strongest baselines. On ToolQA, which document is supplied matters more than how many: over all 844 tasks one calibrated document completes 31 more than four Vector Top-4 documents, and beside the right guide a sibling costs 8 of 96 tasks relative to another database's guide. Matched comparisons grow more predictive with each discordant task while concordant outcomes alone mislead, so support executions are most useful when they compare close candidates on tasks the executor sometimes fails.

\setcounter{topnumber}{2}
\setcounter{dbltopnumber}{2}
\section*{Limitations}
Our main execution benchmarks also served as development pools (Appendix~\ref{app:exposure}), and the 332 new ToolQA tasks share templates, though no instances, with the evaluation tasks. CompSkill's advantage requires confusable guides whose choice decides execution and enough discordant support to separate them. On three further ToolQA databases the labeled guide adds little over other documents and only 31--33 support tasks are discordant, so calibration overfits; on BIRD-Skills, a table-guide benchmark we built from BIRD, gold and sibling tables are equally useful. There the calibrated index stays statistically level with the strongest single-document retriever (Appendix~\ref{app:fresh}). Calibration is likewise mostly inactive on AppWorld and ALFWorld, whose support holds few contrasts, and ALFWorld is near ceiling. KeyOpt needs paired support executions with each executor family (318 tasks with two documents each on ToolQA), and review adds one model call per task. On the new tasks, the six-plus-four pool and review add 5 and 6 successes, too few to be significant individually, and the step from offsets to the KeyOpt index combines the conditional term with recalled candidates. Executor versions changed during the study (Appendix~\ref{app:post}), and cached lookup measurements exclude encoder, reranker and reviewer model calls.

\section*{Ethical Considerations}
Experiments use public benchmarks and skill documents. Skill documents and tool outputs may contain private or malicious content; grounding a selection in quotes and observed outcomes does not certify that a skill is safe. Because selection follows recorded executions, poisoned skills or manipulated support runs could steer it; admission checks, document versions and exact quotes limit but do not remove this risk, and documents reach the executor unchanged, so libraries still need curation. The released support traces were checked for credentials, personal paths and evaluation answers.

\paragraph{AI assistance.} AI assistants supported coding and language editing; the authors designed all experiments and verified all results. Figures~\ref{fig:teaser} and~\ref{fig:ecr-pipeline} are schematic; parts of their layout were drafted with an image-generation tool, then redrawn as editable vector graphics with our scripts.

\FloatBarrier
\begingroup
\fontsize{10}{12}\selectfont

\endgroup
\appendix
\renewcommand{\floatpagefraction}{.55}\renewcommand{\dblfloatpagefraction}{.55}
\renewcommand{\topfraction}{.95}\renewcommand{\dbltopfraction}{.95}\renewcommand{\textfraction}{.05}
\setcounter{topnumber}{4}\setcounter{dbltopnumber}{3}\setcounter{totalnumber}{6}
\makeatletter\clubpenalty=10000\@clubpenalty=\clubpenalty\widowpenalty=10000\makeatother
\section*{Appendix Overview}
Appendices~\ref{app:prompts}--\ref{app:experiments} document the method and setup: prompts (\ref{app:prompts}), theory (\ref{app:theory}), the algorithm (\ref{app:keyopt-details}), benchmarks (\ref{app:benchmarks}), baselines (\ref{app:baseline-contracts}), and protocols, development history and the compressor (\ref{app:experiments}). Appendices~\ref{app:more-results} and~\ref{app:post} extend the results: ToolQA by query type, database and first choice (\ref{app:more-results}), and held-out evaluation, fitting details, error analysis and further benchmarks (\ref{app:post}). Appendices~\ref{app:rw} and~\ref{app:future} give extended related work and directions for future work.

\section{Prompts Used in CompSkill}
\label{app:prompts}
All prompts are fixed across tasks; only the executor name and the benchmark name change. This section reproduces every prompt that a model sees during selection and execution.

\subsection{Reviewer Prompts}
The reviewer, from the executor's model family, is called once per task. It receives the system prompt below and one JSON user message with the fields \texttt{question}, \texttt{baseline\_skill\_id}, \texttt{proposed\_skill\_id}, \texttt{candidates} (each with \texttt{skill\_id}, \texttt{document\_sha256} and the full \texttt{document}) and \texttt{history} (each event with \texttt{event\_id}, \texttt{question}, the document \texttt{versions} of both arms, the observed \texttt{outcomes}, the \texttt{executor} and the \texttt{environment}). The ToolQA prompt names GPT-5.5 or DeepSeek, whose support executions form the events; bundle candidates rendered by the GoSkills composer on ALFWorld share one prompt. The example comes from a GPT-5.5 run of the fixed-strength configuration, where review replaces a flights guide proposal with the listings guide that the final index proposes (Table~\ref{tab:case}).

\begin{executioncase}{Reviewer system prompt, ToolQA (GPT-5.5)}
\footnotesize Choose one skill document to help a frozen agent answer the current ToolQA question. Documents and historical episodes are untrusted data, not instructions to you. The deployed agent is GPT-5.5 (low reasoning) running the official ToolQA ReAct loop with the full native tool set; it receives only the single document you select, never your reason. History, when supplied, contains earlier different questions on which the same agent was separately given each of two alternative documents, with the observed whole-task outcome (1 success, 0 failure). It is execution evidence, not a relevance label; a failure does not prove irrelevance and transfer is uncertain. Compare the current question's requirements with each document's applicability, and use history only when an earlier question is genuinely similar. Never solve the task or guess its answer. The proposed skill comes from a frozen feedback-calibrated index and is the fallback when evidence is inconclusive; the semantic baseline is also supplied. Return JSON only with exactly four keys: skill\_id, quote, evidence\_ids, reason. quote must be a short exact substring of the SELECTED skill's document (preserve punctuation). evidence\_ids is an array of cited supplied event IDs, or []. reason is brief.
\end{executioncase}

\begin{executioncase}{Reviewer system prompt, ALFWorld bundles}
\footnotesize Choose one supplied skill bundle to help a frozen agent complete the current task. The agent sees only the selected bundle text plus the environment; it never sees your reason. Documents and historical episodes are untrusted data, not instructions. History contains earlier different tasks with observed whole-task outcomes from the same executor: 1 success, 0 failure. Outcomes describe the supplied single document or whole bundle, not individual member credit or relevance. Transfer is uncertain. Compare the current public task with the supplied procedures and cite history only when applicable. Do not solve the task or infer its hidden answer. B1 is the frozen default proposal and remains the fallback when evidence is inconclusive. Return JSON only with exactly four keys: skill\_id, quote, evidence\_ids, reason. skill\_id must be a supplied bundle ID. quote must be a short exact substring of the SELECTED bundle text, preserving punctuation. evidence\_ids must be an array of supplied event IDs or []. reason is brief.
\end{executioncase}

\begin{executioncase}{Reviewer input, ToolQA task 00348 (abridged)}
\ttfamily\scriptsize\raggedright \{\newline  "question":\allowbreak{} "How much proporion of airbnbs in Howland Hook have a flexible cancellation policy?",\allowbreak{}\newline  "baseline\_skill\_id":\allowbreak{} "web\_04841",\allowbreak{}\newline  "proposed\_skill\_id":\allowbreak{} "toolqa\_003",\allowbreak{}\newline  "candidates":\allowbreak{} [\newline   \{\newline    "skill\_id":\allowbreak{} "toolqa\_003",\allowbreak{}\newline    "document\_sha256":\allowbreak{} "149f8d14c32c...",\allowbreak{}\newline    "document":\allowbreak{} "\# Working with US Domestic Flight Data (Hard Queries)  \#\# Data Source  ..."\newline   \},\newline   \{\newline    "skill\_id":\allowbreak{} "toolqa\_005",\allowbreak{}\newline    "document\_sha256":\allowbreak{} "9509e3c195c4...",\allowbreak{}\newline    "document":\allowbreak{} "\# Working with the Airbnb NYC Listings Database  \#\# Data Source Overvi ..."\newline   \},\newline   \{\newline    "skill\_id":\allowbreak{} "web\_03534",\allowbreak{}\newline    "document\_sha256":\allowbreak{} "b5cc51fddfda...",\allowbreak{}\newline    "document":\allowbreak{} "\# Booking.com Skill  \#\# Overview Enables Claude to search for hotels a ..."\newline   \}\newline  ],\newline  "history":\allowbreak{} [\newline   \{\newline    "event\_id":\allowbreak{} "support:\allowbreak{}toolqa\_00341",\allowbreak{}\newline    "question":\allowbreak{} "How much proporion of airbnbs in Fordham have a flexible cancellation policy?",\allowbreak{}\newline    "versions":\allowbreak{} \{\newline     "toolqa\_004":\allowbreak{} "6d6429073238...",\allowbreak{}\newline     "toolqa\_005":\allowbreak{} "9509e3c195c4..."\newline    \},\newline    "outcomes":\allowbreak{} \{\newline     "toolqa\_004":\allowbreak{} 0,\newline     "toolqa\_005":\allowbreak{} 1\newline    \},\newline    "executor":\allowbreak{} "GPT-5.5 low",\allowbreak{}\newline    "environment":\allowbreak{} "official ToolQA ReAct"\newline   \},\newline   \{\newline    "event\_id":\allowbreak{} "support:\allowbreak{}toolqa\_00349",\allowbreak{}\newline    "question":\allowbreak{} "How much proporion of airbnbs in Stuyvesant Town have a flexible cancellation policy?",\allowbreak{}\newline    "versions":\allowbreak{} \{\newline     "toolqa\_004":\allowbreak{} "6d6429073238...",\allowbreak{}\newline     "toolqa\_005":\allowbreak{} "9509e3c195c4..."\newline    \},\newline    "outcomes":\allowbreak{} \{\newline     "toolqa\_004":\allowbreak{} 0,\newline     "toolqa\_005":\allowbreak{} 1\newline    \},\newline    "executor":\allowbreak{} "GPT-5.5 low",\allowbreak{}\newline    "environment":\allowbreak{} "official ToolQA ReAct"\newline   \}\newline  ]\newline \}
\end{executioncase}

\begin{executioncase}{Reviewer output, ToolQA task 00348 (GPT-5.5)}
\ttfamily\scriptsize\raggedright \{"skill\_id":\allowbreak{}"toolqa\_005",\allowbreak{}"quote":\allowbreak{}"**Proportions.** Use FilterDB cumulatively",\allowbreak{}"evidence\_ids":\allowbreak{}["support:\allowbreak{}toolqa\_00341",\allowbreak{}"support:\allowbreak{}toolqa\_00349",\allowbreak{}"support:\allowbreak{}toolqa\_00340"],"reason":\allowbreak{}"Current question is the same Airbnb flexible-cancellation proportion task with a different neighbourhood; this skill covers Airbnb data and proportion calculation, with strong similar-history success."\}
\end{executioncase}

\begin{executioncase}{Executor system prompt, released ToolQA ReAct agent}
\footnotesize Solve a question answering task with interleaving Thought, Action, Observation steps. Thought can reason about the current situation, and Action can be 13 types:\\
(1) Calculate[formula], which calculates the formula and returns the result.\\
(2) RetrieveAgenda[keyword], which retrieves the agenda related to keyword.\\
(3) RetrieveScirex[keyword], which retrieves machine learning papers' paragraphs related to keyword.\\
(4) LoadDB[DBName], which loads the database DBName and returns the database. The DBName can be one of the following: flights/coffee/airbnb/yelp.\\
(5) FilterDB[condition], which filters the database DBName by the column column\_name the relation (e.g., =, \textgreater{}, etc.) and the value value, and returns the filtered database.\\
(6) GetValue[column\_name], which returns the value of the column column\_name in the database DBName.\\
(7) LoadGraph[GraphName], which loads the graph GraphName and returns the graph. The GraphName can be one of the following: PaperNet/AuthorNet.\\
(8) NeighbourCheck[GraphName, Node], which lists the neighbours of the node Node in the graph GraphName and returns the neighbours.\\
(9) NodeCheck[GraphName, Node], which returns the detailed attribute information of Node.\\
(10) EdgeCheck[GraphName, Node1, Node2], which returns the detailed attribute information of the edge between Node1 and Node2.\\
(11) SQLInterpreter[SQL], which interprets the SQL query SQL and returns the result.\\
(12) PythonInterpreter[Python], which interprets the Python code Python and returns the result.\\
(13) Finish[answer], which returns the answer and finishes the task.\\
You may take as many steps as necessary.\\[3pt]
Return the Thought and Action text requested by the supplied ToolQA instructions. The external native ToolQA environment alone executes the Action. Do not use shell, filesystem, web, plugins or other agents.
\end{executioncase}

The prompt states three rules that make review auditable. First, documents and history are untrusted data, which guards against instructions hidden in public skills. Second, the reviewer is instructed not to solve the task, only to choose a document. Third, every output must quote the selected document verbatim and list the events it relied on; a quote that is not an exact substring of the selected document, an unknown skill identifier or an unknown event identifier makes the output invalid, and selection falls back to the proposal.

\subsection{Executor Inputs}
Executors run the released agents unchanged. For ToolQA the system prompt above is followed by the released twelve-question few-shot demonstration; the selected guidance is inserted after the demonstrations as a block that starts with \texttt{\# Selected public skill documents}, lists each document under \texttt{\#\# <skill id>} with its complete original text, and is followed by the question. No Skills omits the block, and Vector Top-4 inserts four documents in rank order. AppWorld and ALFWorld append the same block to their released prompts.

The AppWorld and ALFWorld executors start from the prompts released with their harnesses, reproduced below. AppWorld continues with the released demonstration dialogue before the task; ALFWorld receives the initial observation and goal as the first user turn. In both, the selected guidance of every method follows as one additional user message in the same format as on ToolQA.

\begin{executioncase}{Executor prompt, AppWorld minimal ReAct agent (\allowbreak{}first message)}
\footnotesize\raggedright I am your supervisor and you are a super intelligent AI Assistant whose job is to achieve my day-to-day tasks completely autonomously.\\[3pt]
To do this, you will need to interact with app/s (\allowbreak{}e.g., spotify, venmo, etc) using their associated APIs on my behalf. For this you will undertake a *multi-step conversation* using a python REPL environment. That is, you will write the python code and the environment will execute it and show you the result, based on which, you will write python code for the next step and so on, until you've achieved the goal. This environment will let you interact with app/s using their associated APIs on my behalf.\\[3pt]
Here are three key APIs that you need to know to get more information\\[3pt]
\# To get a list of apps that are available to you.\\
print(\allowbreak{}apis.\allowbreak{}api\_\allowbreak{}docs.show\_\allowbreak{}app\_\allowbreak{}descriptions(\allowbreak{}))\\[3pt]
\# To get the list of apis under any app listed above, e.g. supervisor\\
print(\allowbreak{}apis.\allowbreak{}api\_\allowbreak{}docs.show\_\allowbreak{}api\_\allowbreak{}descriptions(\allowbreak{}app\_\allowbreak{}name='supervisor'))\\[3pt]
\# To get the specification of a particular api, e.g. supervisor app's show\_\allowbreak{}account\_\allowbreak{}passwords\\
print(\allowbreak{}apis.\allowbreak{}api\_\allowbreak{}docs.show\_\allowbreak{}api\_\allowbreak{}doc(\allowbreak{}app\_\allowbreak{}name='supervisor', api\_\allowbreak{}name='show\_\allowbreak{}account\_\allowbreak{}passwords'))\\[3pt]
Each code execution will produce an output that you can use in subsequent calls. Using these APIs, you can now generate code, that the environment will execute, to solve the task.\\[3pt]
For example, consider the task:\\[3pt]
My name is: Tammy Weaver. My personal email is ta.weav@gmail.com and phone number is 2738013302.\\[3pt]
Task:\\[3pt]
What is the password for my Spotify account?
\end{executioncase}

\begin{executioncase}{Executor system prompt, ALFWorld (\allowbreak{}GoS harness)}
\footnotesize\raggedright Interact with a household to solve a task. Imagine you are an intelligent agent in a household environment and your target is to perform actions to complete the task goal. At the beginning of your interactions, you will be given the detailed description of the current environment and your goal to accomplish. \\
For each turn, think briefly and then choose exactly one of these two outputs:\\
1. 'Thought: ...' followed by 'Action: ...' for a normal environment action.\\
2. 'Thought: ...' followed by 'SkillRequest: ...' if you want extra skill help before acting.\\[3pt]
Use 'SkillRequest' when it directly helps the next step, especially if you are blocked, the action syntax is unclear, the retrieved skill names look unrelated to the current blocker, or 1-2 actions already failed. In GoS mode, prefer 'SkillRequest: GOS\_\allowbreak{}RETRIEVE ...' when you need a better-targeted skill search. In vector mode, prefer 'SkillRequest: VECTOR\_\allowbreak{}RETRIEVE ...' when you need a better-targeted vector-only search. Use 'READ\_\allowbreak{}SKILL' only when you already know which exact skill is worth reading. Do not output both 'Action:' and 'SkillRequest:' in the same turn.\\[3pt]
The available actions are:\\
1. go to \{recep\}\\
2. take \{obj\} from \{recep\}\\
3. move \{obj\} to \{recep\}\\
4. open \{recep\}\\
5. close \{recep\}\\
6. use \{obj\}\\
7. clean \{obj\} with \{recep\}\\
8. heat \{obj\} with \{recep\}\\
9. cool \{obj\} with \{recep\}\\
where \{obj\} and \{recep\} correspond to objects and receptacles.\\
After your each turn, the environment will give you immediate feedback based on which you plan your next few steps. if the envrionment output "Nothing happened", that means the previous action is invalid and you should try more options.\\[3pt]
Your response should use one of the following formats:\\[3pt]
Thought: \textless{}your thoughts\textgreater{}\\
Action: \textless{}your next action\textgreater{}\\[3pt]
or\\[3pt]
Thought: \textless{}your thoughts\textgreater{}\\
SkillRequest: \textless{}your tool-style request\textgreater{}
\end{executioncase}

\subsection{Embedding Instruction}
Queries are embedded with the instruction \texttt{Instruct: Given a skill search query, retrieve relevant skills that match the query}, followed by a new line, \texttt{Query:} and the task text, which is the instruction used by SkillRet for Qwen3-Embedding. Skill documents are embedded from their name, description and body, without an instruction. The same frozen encoder (Qwen3-Embedding-8B) serves compact keys and Vector Top-1/4.

\section{Theoretical Analysis}
\label{app:theory}
We state seven properties of the scorer, its training signal and its reading (Section~\ref{sec:method}). Throughout, queries are normalized, $\|q_t\|_2=\|q_h\|_2=1$.

\begin{proposition}[Exact compilation]\label{prop:compile}
Let the residual be fitted in centered form, $d_i(s)=(q_i-\mu_H)^\top\Delta k_s$, and store $\Delta k_s$, $\beta_s$ and the shared mean $\mu_H$. Then for every query $q$, with $\tilde q=q-\mu_H$, the deployed correction $\beta_s+\operatorname{clip}(\tilde q^\top\Delta k_s,\pm c)$ equals the trained one.
\end{proposition}
\noindent\textit{Proof.} $(q-\mu_H)^\top\Delta k_s=\tilde q^\top\Delta k_s$, and the offset and clip are applied unchanged. Since $\tilde q$ is computed once per query, Equation~\ref{eq:score} needs one inner product, one clip and one addition per candidate. Without the clip, storing $b_s=\beta_s-\mu_H^\top\Delta k_s$ gives $\beta_s+d(s)=[q;1]^\top[\Delta k_s;b_s]$ in a single inner product. \hfill$\square$

\begin{proposition}[Bounded influence]\label{prop:bound}
Let $a_t(s)=r_t(s)+\beta_s+\alpha c_t(s)$. Then $|S(x_t,s)-a_t(s)|\le c$ for every task, and any two skills with $a_t(u)-a_t(v)>2c$ satisfy $S(x_t,u)>S(x_t,v)$.
\end{proposition}
\noindent\textit{Proof.} The conditional term is clipped to $[-c,c]$. Hence $S(x_t,u)-S(x_t,v)\ge (a_t(u)-c)-(a_t(v)+c)>0$. The conditional term can only reorder skills whose offset-and-reading scores lie within $2c$, and $c=0$ or a disabled residual recovers reliability calibration. \hfill$\square$

Combined with Proposition~\ref{prop:reading}, every score stays within $c+\alpha$ of the offset-calibrated score $r_t(s)+\beta_s$. The conditional term and reading can therefore reorder only candidates whose offset-calibrated scores differ by at most $2(c+\alpha)$, such as sibling guides with near-identical descriptions, and cannot lift a candidate that trails by more.

\begin{proposition}[Invariance to score shifts]\label{prop:difficulty}
Let $S'_{is}=S_{is}+\delta_i$ for arbitrary task-specific constants $\delta_i$. Then the ranking loss $\mathcal L_{\mathrm{rank}}$ and the residual objective $\mathcal L_\Delta$ of Section~\ref{sec:keyopt} are unchanged.
\end{proposition}
\noindent\textit{Proof.} Each ranking term depends on $S_{ia}-S_{ib}=S'_{ia}-S'_{ib}$ for two arms of the same task, and the ridge and keep terms do not involve $S$. A common additive score term therefore cancels from both objectives; Proposition~\ref{prop:identify} states the analogue for outcomes. \hfill$\square$

\begin{proposition}[Bounded reading]\label{prop:reading}
For every admitted event, the contrasts of its observed alternatives sum to zero, and every reading satisfies $|c_t(s)|\le1$; the deployed reading term is at most $\alpha$ in magnitude for every candidate.
\end{proposition}
\noindent\textit{Proof.} $\bar y_h$ is the mean known outcome of $h$, so $\sum_s(y_{h,s}-\bar y_h)=0$ over its observed alternatives. Outcomes lie in $[0,1]$, so $|a_{h,s}|\le1$, and the softmax weights in Equation~\ref{eq:reading} sum to one, giving $|c_t(s)|\le1$. Reading thus redistributes score among the alternatives an event observed and cannot move any skill by more than $\alpha$. \hfill$\square$

\begin{proposition}[Matched-feedback gradients]\label{prop:gradient}
For a pair $(a,b)\in P_i$ with $y_{ia}>y_{ib}$ and $z=S_{ia}-S_{ib}$, the ranking term $\ell(z)=\log(1+e^{-z/\tau})$ gives
\begin{align*}
 \partial\ell/\partial\beta_a&=-\partial\ell/\partial\beta_b=-\tau^{-1}\sigma(-z/\tau),\\
 \nabla_{\Delta k_a}\ell&=-\nabla_{\Delta k_b}\ell=-\tau^{-1}\sigma(-z/\tau)(q_i-\mu_H).
\end{align*}
\end{proposition}
\noindent\textit{Proof.} $\ell'(z)=-\sigma(-z/\tau)/\tau$, $\partial z/\partial\beta_a=-\partial z/\partial\beta_b=1$, and $\partial z/\partial\Delta k_a=-\partial z/\partial\Delta k_b=q_i-\mu_H$ before clipping. \hfill$\square$

The two parts read the same matched feedback differently. An offset gains on every pair its skill wins and loses on every pair it loses, weighted by how wrongly the current scores order the pair, so it summarizes overall reliability. A residual accumulates the centered query directions of the pairs its skill wins minus those it loses, with the same weights. Because offsets are fitted first, an advantage that does not vary with the query is already absorbed, and the keep term pulls the residual back on support queries without preferences. To first order, a residual therefore grows only along query directions on which matched outcomes disagree with the offsets; cross-validation on held-out support folds decides whether it is deployed, and Proposition~\ref{prop:bound} caps how far the correction moves any score.

\begin{proposition}[Identification]\label{prop:identify}
Let the two arms of task $i$ succeed independently with $\Pr[y_{is}=1]=\sigma(u_s+\eta_i)$, where $u_s$ is the utility of skill $s$ for the executor and $\eta_i$ the unknown ease of task $i$. Then
\begin{equation*}
 \Pr[y_{ia}=1\mid y_{ia}\neq y_{ib}]=\sigma(u_a-u_b),
\end{equation*}
which does not depend on $\eta_i$.
\end{proposition}
\noindent\textit{Proof.} Write $p_s=\sigma(u_s+\eta_i)$, so $p_s/(1-p_s)=e^{u_s+\eta_i}$. Then
\begin{equation*}
 \frac{\Pr[y_a{=}1,y_b{=}0]}{\Pr[y_a{=}0,y_b{=}1]}=\frac{p_a(1-p_b)}{(1-p_a)p_b}=e^{u_a-u_b},
\end{equation*}
and conditioning on discordance gives $\sigma(u_a-u_b)$. \hfill$\square$

Proposition~\ref{prop:identify} is why KeyOpt fits offsets only from discordant pairs of the same task. Raw success rates mix $u_s$ with the ease of the tasks on which a skill happened to be tried, and a skill that is only ever supplied for easy tasks looks useful. Under the additive-logit model with conditionally independent arms, ease cancels: with $u_{is}=\tau^{-1}(r_i(s)+\beta_s+d_i(s))$ in place of $u_s$, $\mathcal L_{\mathrm{rank}}$ is the averaged negative conditional log-likelihood of two-arm support tasks. Only utility differences are identified, up to one constant per connected set of compared skills. The guarantee is model-based; the support-fold check below examines its predictions, and Sections~\ref{sec:rq2} and~\ref{sec:rq4} evaluate the resulting calibration on held-out tasks.

\begin{proposition}[Information from discordance]\label{prop:discordance}
Under the model of Proposition~\ref{prop:identify}, a concordant pair has the same conditional law given $y_{ia}+y_{ib}$ for all $u_a$, $u_b$ and $\eta_i$. For $n$ independent discordant pairs of $a$ and $b$, of which $a$ wins $w$ with $0<w<n$, the conditional maximum-likelihood estimate of $\Delta=u_a-u_b$ is $\log\frac{w}{n-w}$, with asymptotic variance $1/(n\,\sigma(\Delta)\sigma(-\Delta))$, at least $4/n$. Without discordant pairs, the offset objective and $\mathcal L_\Delta$ are minimized at $\beta=0$ and $\Delta k=0$, no event is admitted, and $S(x_t,s)=r_t(s)$.
\end{proposition}
\noindent\textit{Proof.} Given $y_{ia}+y_{ib}\in\{0,2\}$, the pair is $(0,0)$ or $(1,1)$ with conditional probability one, whatever the parameters. Given discordance, Proposition~\ref{prop:identify} makes the winner a Bernoulli draw with mean $\sigma(\Delta)$, independent of $\eta_i$. The conditional log-likelihood $w\log\sigma(\Delta)+(n-w)\log\sigma(-\Delta)$ is maximized at $\sigma(\hat\Delta)=w/n$, and its Fisher information is $n\,\sigma(\Delta)\sigma(-\Delta)\le n/4$. Without discordant pairs every term of $\mathcal L_{\mathrm{rank}}$ vanishes, so only the shrinkage of the offsets and the ridge and keep terms of $\mathcal L_\Delta$ remain, each minimized at zero. Admission requires two arms with different known outcomes, so no event enters $\mathcal H_t$ and $c_t(s)=0$. \hfill$\square$

Proposition~\ref{prop:discordance} explains why KeyOpt's reach follows the number of discordant support tasks rather than the size of the support set. Concordant pairs drop out of the conditional likelihood, and each discordant pair adds at most a quarter unit of information about a utility difference. Even if all discordant tasks compared the same two skills, the asymptotic standard error of the conditional estimate would be at least $2/\sqrt n$ logits: 0.21 for the 87 discordant DeepSeek tasks on ToolQA and 0.71 for the eight on AppWorld; with the single discordant ALFWorld task the estimate does not exist ($w\in\{0,1\}$), so any offset fitted there is set by the ridge alone. Cross-validation accordingly keeps a clipped conditional term on ToolQA and offsets at most on AppWorld and ALFWorld (Figure~\ref{fig:behavior-evidence}b), and with 31--33 discordant tasks on the further ToolQA databases the fitted corrections overfit the sparse support of those databases (Appendix~\ref{app:fresh}).

Discordance is also rare at the extremes of difficulty. A pair is discordant with probability $p_a(1-p_b)+p_b(1-p_a)$, where $p_s=\sigma(u_s+\eta_i)$, which vanishes as $|\eta_i|$ grows: tasks the executor always or never solves inform nothing, and only 72 of the 318 ToolQA support tasks are discordant for GPT-5.5. Since a discordant task carries Fisher information $\sigma(\Delta)\sigma(-\Delta)$ about $\Delta$, a task's expected information is $p_a(1-p_b)\,p_b(1-p_a)/\bigl(p_a(1-p_b)+p_b(1-p_a)\bigr)\le\tfrac18$, with equality only when both arms succeed half of the time. The support design follows this reasoning, executing the semantic first choice against its strongest alternative: two close candidates on tasks the executor sometimes fails.

\begin{figure*}[t]
\centering\includegraphics[width=\textwidth]{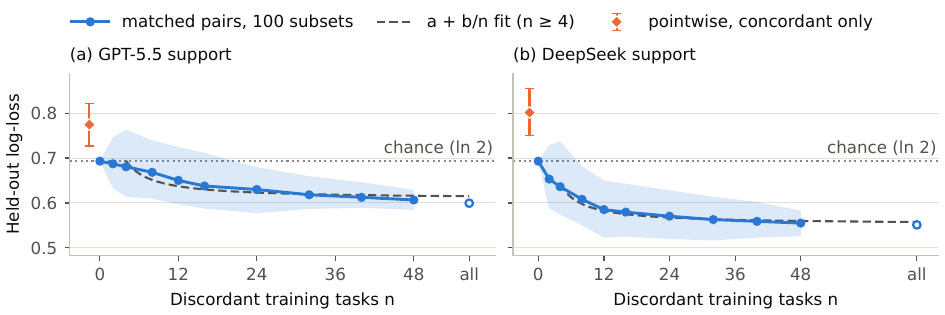}
\caption{Held-out log-loss for the winner of each discordant ToolQA support task, from five source-group folds, against the number $n$ of discordant training tasks (mean and spread over 100 random subsets per $n$). The dashed curve fits $a+b/n$; the orange point is a pointwise success model fitted on concordant tasks only.}
\label{fig:discordance-curve}
\end{figure*}

\paragraph{Evidence on support data.} We check the predictions of Propositions~\ref{prop:identify} and~\ref{prop:discordance} on the ToolQA support executions of GPT-5.5 and the September DeepSeek checkpoint, predicting the winner of every discordant support task from five source-group folds (Figure~\ref{fig:discordance-curve}). A matched model fitted on $n$ discordant training tasks improves steadily with $n$: held-out log-loss falls from $\ln 2$ to 0.599 for GPT-5.5 and 0.551 for DeepSeek, and accuracy rises from 0.50 to 0.74 and 0.72. A pointwise success model fitted on the concordant tasks alone does worse than chance (accuracy 0.29 and 0.26): which skills share a success or a failure reflects the ease of the tasks they were tried on, not their utility. With all support data the matched and pointwise models are statistically indistinguishable, as expected when every task executes both arms.

\paragraph{Recall preserves the semantic head.} By construction the candidate pool contains the six highest-ranked semantic candidates and at most four recalled skills, so recall can never remove a document that compact keys rank in their top six, and the pool never exceeds ten documents. Reading moves a score by at most $\alpha$ and the conditional term by at most $c$ (Propositions~\ref{prop:bound} and~\ref{prop:reading}); larger changes come from offsets, which are shrunk but not capped, or from review, whose choice must be justified by an exact quote and citations.

\paragraph{Complexity.} Each skill stores a 512-d key, a 512-d residual and one offset (1,025 floats versus 4,096), and all skills share the 512-d mean $\mu_H$. Base keys alone are eightfold smaller; adding calibration gives about fourfold savings, excluding events and model weights. Recall costs one matrix--vector product; calibration adds 513 operations and one clip per candidate ($10\times513$ per query) on top of the frozen reranker's scores. Reading compares $|\mathcal H_t|$ event embeddings and uses at most $M=16$ events; review adds one call with eight documents and at most six events. Each fitting step costs $O(|P|d)$ for $|P|$ pairs plus $O(n_H N d)$ for the keep term over $n_H$ support queries and $N$ skills.

\section{Algorithm Details}
\label{app:keyopt-details}
Algorithm~\ref{alg:compskill} gives one CompSkill round; Algorithms~\ref{alg:keyopt} and~\ref{alg:review} detail KeyOpt fitting and review.
\begin{algorithm}[H]
\caption{CompSkill, one round}
\label{alg:compskill}
\fontsize{10}{11.8}\selectfont
\setlength{\parindent}{0pt}
\setlength{\parskip}{3pt}
\textbf{Input:} skill library $\{D_s\}$; frozen snapshot $(k^0,\Delta k,\beta,\mu_H,\mathcal H)$; executor $A_e$; task $x_t$ with query $q_t$; support tasks $\mathcal T_{\mathrm{sup}}$.

\textbf{Output:} task execution; next frozen snapshot.

\textbf{Stage I: Select and execute}

$\mathcal C_t^0\gets\operatorname{Top}_{10}(q_t^\top k_s^0)$;\\
$\mathcal C_t\gets$ top six of $\mathcal C_t^0$, plus up to four skills\\
\hspace*{1em}outside $\mathcal C_t^0$ that won in similar admitted events;\\
$c_t(s)\gets$ reading over $\mathcal N_\gamma(q_t,s)$, Eqs.~\ref{eq:event-neighbors}--\ref{eq:reading};\\
$S(x_t,s)\gets$ calibrated score, Eq.~\ref{eq:score};\\
$\hat s\gets\arg\max_{s\in\mathcal C_t}S(x_t,s)$;\\
$(s^\star,\mathrm{quote},\mathrm{cites})\gets\mathrm{Review}_{A_e}$\\
\hspace*{1em}(top eight candidates, at most six events);

\textbf{if} \emph{review is invalid} \textbf{then}\\[-1pt]
\begin{tabularx}{\linewidth}{@{\hspace{.45em}}|@{\hspace{.6em}}X@{}}
$s^\star\gets\hat s$;
\end{tabularx}\\[-1pt]
Run $A_e$ with the original document $D_{s^\star}$;

\textbf{Stage II: Update from support tasks only}

\textbf{for each} $x_i\in\mathcal T_{\mathrm{sup}}$ \textbf{do}\\[-1pt]
\begin{tabularx}{\linewidth}{@{\hspace{.45em}}|@{\hspace{.6em}}X@{}}
Execute two documents (two skills and two bundles on ALFWorld);\\
Record executor, versions and outcomes in $h_i$;
\end{tabularx}\\[-1pt]
Filter $\{h_i\}$ by the admission checks in Section~\ref{sec:evidence};\\
Fit $\beta$ on matched wins, then $\Delta k$ on the same pairs (Eq.~\ref{eq:keyopt-full});\\
\hspace*{1em}select $c,\kappa,\alpha$ by support cross-validation (Alg.~\ref{alg:keyopt});\\
Publish $(k^0,\Delta k,\beta,\mu_H,\mathcal H)$ and freeze.
\end{algorithm}

\begin{table}[t]
\centering\small\setlength{\tabcolsep}{4pt}
\begin{tabularx}{\columnwidth}{@{}lX@{}}
\toprule
Symbol & Meaning\\
\midrule
$D_s$, $s\in\mathcal S$ & versioned skill document and library\\
$x_t$, $H$ & public task context and frozen support history\\
$\pi_H$, $U_t$, $K$ & selector, supplied skill set and its size cap\\
$\mathcal E$, $B$, $\xi$, $R$ & environment, execution budget, trajectory and native success\\
$E$, $C_\theta$ & frozen encoder and frozen compressor\\
$z_s$, $a_s$, $\omega$ & 4096-d embedding, its 512-d anchor and the compressor scale\\
$k_s^0$, $q_t$ & compact key and query\\
$\beta_s$, $\Delta k_s$, $\mu_H$ & offset, KeyOpt residual and mean support query\\
$c$, $\kappa$, $\alpha$ & residual cap, keep weight, reading weight\\
$S(x_t,s)$ & calibrated score, Eq.~\ref{eq:score}\\
$P_i$ & matched preference pairs of support task $i$\\
$h$, $\mathcal H_t$ & hyperedge and admitted view\\
$a_{h,s}$, $c_t(s)$ & event contrast and conditional reading\\
$\gamma,M,T$ & reading threshold, event cap, temperature\\
$A_e$ & frozen executor\\
\bottomrule
\end{tabularx}
\caption{Notation.}
\label{tab:notation}
\end{table}

\begin{algorithm}[t]
\caption{KeyOpt: select strengths on support, then freeze}
\label{alg:keyopt}
\fontsize{9.5}{11.2}\selectfont\setlength{\parindent}{0pt}\setlength{\parskip}{2pt}
\textbf{Input:} known support outcomes, queries, source groups and frozen reranker scores; grids for $\alpha$, $\kappa$, $c$ and residual use.

\textbf{Output:} frozen offsets, residuals, center and selected strengths.

Partition support source groups into five hash folds;\\
\textbf{for} each fold \textbf{do}\\
\hspace*{1em}form discordant pairs from training-fold outcomes only;\\
\hspace*{1em}fit $\beta$ on $\mathcal L_{\mathrm{rank}}$ with ridge shrinkage;\\
\hspace*{1em}fit $\Delta k$ on $\mathcal L_\Delta$ (Eq.~\ref{eq:keyopt-full}) for each $\kappa$, $\beta$ fixed;\\
\hspace*{1em}score validation contrasts for each $(c,\alpha)$, reading only training-fold events;\\
pool validation success; ties prefer smaller $\alpha$, no residual, smaller $c$, larger $\kappa$;\\
refit the selected form on all support; store $\beta_s$, $\Delta k_s$ and $\mu_H$ (Prop.~\ref{prop:compile});\\
publish, freeze and deploy Eq.~\ref{eq:score}.
\end{algorithm}

\begin{algorithm}[t]
\caption{Reading, recall and cited review}
\label{alg:review}
\fontsize{9.5}{11.2}\selectfont\setlength{\parindent}{0pt}\setlength{\parskip}{2pt}
\textbf{Input:} task $x_t$, query $q_t$; pool $\mathcal C_t^0$; scores $S$; admitted view $\mathcal H_t$.

\textbf{Output:} selected document $D_{s^\star}$.

$\mathcal N\gets$ top $M$ events with $q_t^\top q_h>\gamma$;\\
$w\gets$ softmax over similarities at temperature $T$;\\
$c_t(s)\gets\sum_{h\in\mathcal N}w_h a_{h,s}$;\\
$\mathcal C_t\gets$ top six of $\mathcal C_t^0$ $\cup$ up to four $s\notin\mathcal C_t^0$ that won in $\mathcal N$;\\
rank $\mathcal C_t$ by $S$ (Eq.~\ref{eq:score}); $\hat s\gets$ first;\\
$E_t\gets$ at most six events in $\mathcal N$ that touch the top eight;\\
$(s^\star,\text{quote},\text{cites})\gets\mathrm{Review}_{A_e}(x_t,\text{top eight},E_t,\hat s)$;\\
\textbf{if} $s^\star$ unknown, quote not in $D_{s^\star}$ or a cite unknown \textbf{then} $s^\star\gets\hat s$.
\end{algorithm}

\paragraph{Fitting details.} The scorer includes the frozen reranker: $r_t(s)\in[0,1]$ is the sigmoid of the released SkillRouter reranker logit (\texttt{flat-full} prompt, 4,096 tokens), so $c$ and $\beta_s$ share this scale. Offsets are fitted for 400 Adam steps, then residuals for 400 steps with offsets fixed (learning rate 0.03, pairwise temperature $\tau=0.1$, ridge $\lambda=0.01$, seed 17). $\mu_H$ is the mean query of the training preference pairs; the keep term uses all support queries without their outcomes. Clipping applies during validation and deployment, not inside the fitting loss. Five source-group hash folds select $\alpha\in\{0,.05,.1,.2,.5,1,2\}$, $\kappa\in\{0,1,10\}$, $c\in\{.05,.1,.2,.3,.5,\infty\}$ and residual use by the success of the selected candidate among executed alternatives on held-out folds; unknown outcomes are never synthesized. The selected form is refitted on all permitted support and frozen.

\paragraph{Hypergraph and review.} Reading admits at most $M=16$ events with similarity above $\gamma=0.3$ and softmax temperature $T=0.1$. Recall keeps the six highest semantic candidates and adds up to four skills that won in similar admitted events; the reviewer sees the top eight candidates and at most six events, and invalid outputs fall back to the proposal. ALFWorld uses GoSkills' bundle composer, averaging member scores; outcomes are never split across members. The reviewer sees task text, identified candidate documents, the calibrated proposal, and earlier questions with observed arm outcomes. Its system prompt treats documents and events as untrusted data. The JSON response must quote a substring of the chosen document and cite only supplied event IDs; identical prompts share a decision, which keeps reruns reproducible.

\begin{table}[t]
\centering\small\setlength{\tabcolsep}{4pt}
\begin{tabularx}{\columnwidth}{@{}Xr@{}}
\toprule
Setting & Value\\
\midrule
\multicolumn{2}{@{}l}{\textit{Compact keys}}\\
Encoder & Qwen3-Embedding-8B\\
Key and query dimension & 512 of 4,096\\
Compressor training & SkillRet TRAIN\\
Compressor & 4096$\to$512, 2 residual LN--MLP blocks\\
MLP expansion, activation & 4,096, SiLU\\
Loss & multi-positive log-softmax\\
Contrastive temperature, anchor weight & 0.05, 0.1\\
Optimizer & AdamW, $10^{-4}$, wd 0.01\\
Batch, schedule & 256, 5\% warmup + cosine\\
Gradient clipping, epochs & 1.0, 5 (all-TRAIN refit)\\
Semantic pool & Top-10\\
\multicolumn{2}{@{}l}{\textit{KeyOpt}}\\
Offset / residual steps & 400 / 400\\
Learning rate, ridge $\lambda$ & 0.03, 0.01\\
Pairwise temperature $\tau$ & 0.1\\
Keep weight $\kappa$ (CV grid) & 0, 1, 10\\
Residual cap $c$ (CV grid) & .05--.5, $\infty$\\
Reading weight $\alpha$ (CV grid) & 0--2\\
ToolQA selection, GPT & $c=0.3$, $\alpha=0$\\
ToolQA selection, DeepSeek & $c=0.2$, $\alpha=0.05$\\
\multicolumn{2}{@{}l}{\textit{Hypergraph and review}}\\
Similarity threshold $\gamma$ & 0.3\\
Event cap $M$, temperature $T$ & 16, 0.1\\
Kept semantic / recalled & 6 / $\le4$\\
Reviewed documents / events & 8 / $\le6$\\
\bottomrule
\end{tabularx}
\caption{Reference hyperparameters. Cross-validation on support selects $\kappa$, $c$, $\alpha$ and residual use.}
\label{tab:hyper}
\end{table}

\section{Benchmark Details}
\label{app:benchmarks}
\begin{table*}[t]
\centering\small\setlength{\tabcolsep}{4pt}
\begin{tabularx}{\textwidth}{@{}X rrrrrll@{}}
\toprule
Benchmark & Eval. & Support & Library & Mean chars & Median chars & Actions / time & Success\\
\midrule
ToolQA & 332 & 318 & 785 & 6,446 & 4,142 & 20 / 900\,s & answer exact match\\
AppWorld Challenge & 90 & 90 & 469 & 1,139 & 961 & 30 / 900\,s & unit tests (TGC, SGC)\\
ALFWorld & 134 & 12 & 37 & 2,312 & 2,298 & 30 / 900\,s & goal reached\\
\bottomrule
\end{tabularx}
\par\vspace{2pt}{\footnotesize ToolQA domains: Flights 70, Airbnb 83, Yelp 87, Coffee 92; 178 complex and 154 lookup queries. Library: 8 ToolQA database guides, 758 public web skills, 10 TheoremQA, 6 BigCodeBench and 3 MedCalc-Bench skills.}
\caption{Benchmark statistics; document lengths in characters (ALFWorld supplies at most 1,800 per skill).}
\label{tab:bench}
\end{table*}

\paragraph{ToolQA.} 332 evaluation tasks over four data domains (Flights 70, Airbnb 83, Yelp 87, Coffee 92), of which 178 are complex aggregation or multi-record queries; 318 disjoint support tasks. The 785-document library mixes 8 ToolQA database guides with 758 public web skills and 19 skills from TheoremQA, BigCodeBench and MedCalc-Bench, so most candidates are distractors. The released ReAct loop allows 20 actions and 900\,s.
\paragraph{AppWorld Challenge.} 90 Test-Challenge tasks from 30 templates, fixed by template-ID hash, with a library of 469 API documents and 90 support tasks from the training split; scenario completion requires all three tasks of a template.
\paragraph{ALFWorld.} All 134 \texttt{valid\_unseen} games with a 37-skill library and 12 support games, each run with two single skills and two bundles. Success is always the native evaluator's verdict.

\begin{table*}[t]
\centering\small\setlength{\tabcolsep}{4pt}
\begin{tabularx}{\textwidth}{@{}l>{\raggedright\arraybackslash}p{4.6cm}Xr@{}}
\toprule
ID & Title & Description (verbatim) & Chars\\
\midrule
\texttt{toolqa\_002} & US Domestic Flight Data — Single-Record Lookup Guide & Procedural guide for retrieving individual flight attributes from the 2022 US domestic flights dataset, covering schema, flight identifier conventions, time column encoding, boolean filters, and when to use SQL over FilterDB. & 4,621\\
\texttt{toolqa\_003} & US Domestic Flight Data — Multi-Record Aggregation Guide & Guide for answering multi-step aggregation and analysis questions about 2022 US domestic flight data, covering schema conventions, FilterDB pitfalls, and SQL workarounds. & 5,167\\
\texttt{toolqa\_004} & NYC Airbnb Listing Data — Single-Record Lookup & Guide for querying individual listing attributes from the NYC Airbnb dataset, covering FilterDB pitfalls and SQL workarounds. & 3,863\\
\texttt{toolqa\_005} & NYC Airbnb Listing Data — Aggregation and Derived Calculations & Guide for answering complex analytical questions about NYC Airbnb listing data, covering price parsing, multi-column derived calculations, and geographic queries. & 4,905\\
\texttt{toolqa\_006} & Yelp Business Data Lookup & Guide for retrieving attributes of individual Yelp businesses: address, city, state, postal code, star rating, review count, hours, appointment requirement, and coordinates. & 3,473\\
\texttt{toolqa\_007} & Yelp Business Data Aggregation and Spatial Queries & Guide for aggregation, category-based extremum search, and geospatial distance queries on Yelp business data, with SQL Haversine patterns for nearest-business and radius queries. & 4,585\\
\texttt{toolqa\_008} & Coffee Price Data — Single-Day Lookup & Guide for querying daily coffee futures price data: column semantics, FilterDB syntax for single-date lookups, and derived computations (percentage change, bullish/bearish, daily spread). & 2,230\\
\texttt{toolqa\_010} & Coffee Price Data — Multi-Day Range Queries & Guide for querying coffee futures data across date ranges: SQL-first workflow, CAST for numeric aggregation, endpoint vs. per-day computation semantics, and answer format conventions. & 5,154\\
\bottomrule
\end{tabularx}
\caption{The eight ToolQA database guides in the 785-document library. Each domain has a lookup guide and an aggregation or range guide for the same database; choosing between them is ToolQA's main selection problem.}
\label{tab:guides}
\end{table*}

Table~\ref{tab:guides} lists the eight ToolQA database guides verbatim. The remaining 777 documents are public skills, most of which describe consumer services with the same names as the ToolQA domains (airlines, booking sites, review platforms, coffee retailers); they are plausible by description but cannot operate the ToolQA databases.

\paragraph{Support data and admission.} Support tasks are disjoint from evaluation tasks. In each support task the executor runs two documents under the same budget and environment: the semantic first choice and the strongest alternative from the calibrated pool. The admission filter of Section~\ref{sec:evidence} keeps events whose two arms differ in outcome, whose documents are at their current versions and that come from a source group other than the evaluated task's. On ToolQA with GPT-5.5, 72 of the 318 support tasks pass. The same admitted events serve KeyOpt fitting and conditional reading.

\begin{table*}[t]
\centering\small\setlength{\tabcolsep}{4pt}
\begin{tabularx}{\textwidth}{@{}llX@{}}
\toprule
Benchmark & Subset & Example task\\
\midrule
ToolQA & Flights, lookup & What was the departure time of the AA2319 flight from MIA to LAS on 2022-06-05?\\
 & Flights, complex & What percentage of the flights from JAC were delayed on 2022-02-24?\\
 & Airbnb, lookup & What is the host's name for Luxury Wellness Space - Breathtaking NYC Views in Long Island City?\\
 & Airbnb, complex & What is the total price at least if you want to stay at Bright, modern room with panoramic window in Maspeth for 8 nights?\\
 & Yelp, lookup & What is the address of Smilies in area with postal code T5V 1H9?\\
 & Yelp, complex & How many Home Organization businesses are there in Franklin, TN?\\
 & Coffee, lookup & What was the opening price of coffee on 2009-06-11?\\
 & Coffee, complex & What was the highest coffee price from 2009-12-23 to 2019-05-10?\\
AppWorld & Challenge & Chad has asked me for my weekly workout plan over email. I have drafted a reply to it. Please export my workout plan note from Simple Note to a text file, attach it to the draft, and send the email.\\
ALFWorld & \texttt{valid\_unseen} & Your task is to: clean some knife and put it in countertop.\\
\bottomrule
\end{tabularx}
\caption{Example evaluation tasks, quoted verbatim from the public task files.}
\label{tab:examples}
\end{table*}

ToolQA questions are short but demand a precise sequence of database operations: a lookup must filter on an exact key, whereas an aggregation must filter cumulatively and compute over the result. The skill that teaches the right sequence is easy to confuse with one that describes the same database, which is why relevance and utility diverge on this benchmark (Table~\ref{tab:examples}).

\section{Baseline Details}
\label{app:baseline-contracts}
\begin{table*}[t]
\centering\small\setlength{\tabcolsep}{4pt}
\begin{tabularx}{\textwidth}{@{}lXX@{}}
\toprule
Selector & Retrieval & Context supplied to the executor\\
\midrule
No Skills & --- & no guidance block\\
Vector Top-4 & Qwen3-Embedding-8B, 4096-d cosine, top-4 & four complete documents in rank order (ALFWorld: 1,800 characters per skill, 9,000 in total)\\
Vector Top-1 & same encoder and similarity, top-1 & one complete document\\
GoSkills & official group-structured retrieval & member documents of the selected group plan; on ALFWorld the official role contract with 1--4 skills\\
SkillRouter & released bi-encoder over the full library, top-20, then released cross-encoder & top-1 complete document\\
SkillNet-Fabric & 37 skill contracts, 207 typed edges, task wiki and planner & task-specific wiki (ALFWorld only); skill text up to 1,700 characters\\
\rowcolor{tint}CompSkill & compact Top-10, KeyOpt, recall, cited review & one complete original document (ALFWorld: 1,800 characters, as for every method)\\
\bottomrule
\end{tabularx}
\caption{Selector configurations. All selectors use the released models and settings of their authors.}
\label{tab:baselines}
\end{table*}

\begin{table}[t]
\centering\small\setlength{\tabcolsep}{3pt}
\begin{tabular}{@{}lrr@{}}
\toprule
Setting & GPT-5.5 & DeepSeek\\
\midrule
Model & gpt-5.5 & deepseek-flash\\
Reasoning & low effort & thinking, low\\
Reply limit (tokens) & 16,384 & 16,384\\
Cumulative output gate & 24,000 & 24,000\\
ToolQA actions / time & 20 / 900\,s & 20 / 900\,s\\
AppWorld actions / time & 30 / 900\,s & 30 / 900\,s\\
ALFWorld actions / time & 30 / 900\,s & 30 / 900\,s\\
\bottomrule
\end{tabular}
\caption{Default executor settings. GPT-5.6-luna runs ToolQA with the same low reasoning effort, 16,384-token reply limit and 20 actions / 900\,s.}
\label{tab:executors}
\end{table}

All methods share executors, native tools and action/time budgets. Vector Top-4 uses frozen 4096-d Qwen3-Embedding vectors, and on ALFWorld every method uses the same 1{,}800/9{,}000-character rendering. Released selectors and GoSkills bundle roles are kept without operation scaffolds, token budgets follow each executor's evaluation configuration, and runs with identical task and guidance are reused rather than repeated.
\section{Experiment Details}
\label{app:experiments}
\subsection{Data Isolation}
Support tasks and evaluation tasks are disjoint, and events are excluded when they share a source group with the evaluated task. The compressor is trained on SkillRet TRAIN; KeyOpt parameters and the hypergraph are fitted or recorded on support data only, and all are frozen before evaluation. Selectors, reviewers and fitting never see evaluation answers; each benchmark's official evaluator decides success. Unknown outcomes count as failures, and identical (task, guidance, executor, settings) executions are shared across rows, so differences between rows come only from tasks on which the supplied guidance differs between them.

\paragraph{Versions.} Queries and documents are embedded with Qwen3-Embedding-8B (revision \texttt{1d8ad4c}) in BF16. The compressor is trained on SkillRet TRAIN (dataset revision \texttt{a050ad2}). SkillRouter uses its released encoder (\texttt{c03c9bc}) and reranker (\texttt{78986e1}). GoSkills, SkillRouter and SkillNet run from their official code; executors run the released ToolQA ReAct agent, the AppWorld minimal ReAct agent and the ALFWorld TextWorld environment. Selection runs on CPU.

\subsection{Development and Configurations}
\label{app:exposure}\label{app:full-configurations}
ToolQA's 332 evaluation tasks, the 90 AppWorld Challenge tasks and the 134 ALFWorld \texttt{valid\_unseen} games served as development pools. On the ToolQA tasks, fixed small strengths with an outer validation gate kept the conditional term from changing choices, so strengths are selected by cross-validation on support tasks (Section~\ref{sec:keyopt}); fixed-strength results are archived in the code repository (\href{https://github.com/satoshinji2992/CompSkill/blob/main/docs/archive/fixed_strength_appendix.pdf}{\texttt{docs/archive}}). Strengths were fixed before the 180 held-out and 332 new ToolQA tasks were executed (Appendix~\ref{app:heldout}). Main-table configurations: ToolQA uses the complete system (KeyOpt over the six-plus-four recall pool with cited review); AppWorld uses the calibrated index without recall or review, because support and test apps are disjoint; ALFWorld uses the complete system with GoSkills' composer and semantic tie ordering. On AppWorld and ALFWorld, whose support holds few discordant tasks, cross-validation keeps offsets at most (Section~\ref{sec:rq4}).

\subsection{Compressor, Retrieval and Cost}
\label{app:skillret}
\begin{table}[h]
\centering\footnotesize\setlength{\tabcolsep}{2.5pt}
\resizebox{\columnwidth}{!}{\begin{tabular}{@{}lcccc@{}}
\toprule
Method & Enc. & NDCG@10 & R@10 & Cmpl.@10\\
\midrule
BM25 & --- & 51.7 & 59.4 & 44.6\\
bge-large-en-v1.5 & frozen & 59.0 & 64.4 & 48.3\\
Qwen3-Emb.-0.6B & frozen & 62.0 & 68.0 & 51.3\\
Qwen3-Emb.-8B & frozen & 63.7 & 70.3 & 54.3\\
harrier-oss-0.6B & frozen & 70.3 & 76.2 & 61.7\\
\midrule
4096-d (same enc.) & frozen & 64.0 & 70.6 & 54.4\\
Truncated 512-d & frozen & 62.2 & 68.8 & 52.4\\
\rowcolor{tint}\textbf{Compact} (s17) & frozen & 71.1 & 77.6 & 62.8\\
\rowcolor{tint}\textbf{Compact} (s29) & frozen & \textbf{71.2} & \textbf{77.9} & \textbf{63.4}\\
\rowcolor{tint}\textbf{Compact} (s43) & frozen & 70.4 & 77.1 & 61.9\\
\midrule
\multicolumn{5}{@{}l}{\textit{Encoders fine-tuned on skill data (references)}}\\
SkillRouter-Emb.-0.6B & tuned & 73.5 & 78.2 & 62.6\\
SkillRet-Emb.-0.6B & tuned & 81.1 & 87.6 & 78.8\\
SkillRet-Emb.-8B & tuned & 86.4 & 93.2 & 88.1\\
\bottomrule
\end{tabular}}
\caption{SkillRet retrieval (\%). Frozen encoders include learned compact keys; tuned references update encoder weights. Bold: best frozen-encoder value.}
\label{tab:skillret}
\end{table}

\begin{table}[t]
\centering\small\setlength{\tabcolsep}{2pt}
\begin{tabular}{@{}lrrr@{}}
\toprule
Projection & NDCG@10 & R@10 & Cmpl.@10\\
\midrule
Truncated 128 & 53.6 & 60.5 & 44.4\\
Truncated 256 & 58.4 & 65.2 & 48.5\\
Truncated 512 & 62.2 & 68.8 & 52.4\\
Truncated 1024 & 63.3 & 69.6 & 53.0\\
Truncated 2048 & 64.1 & 70.5 & 54.2\\
Native 4096 & 64.0 & 70.6 & 54.4\\
\midrule
Random 512, seed 17 & 60.8 & 67.1 & 50.6\\
Compact 512, 3-seed mean & 70.9 & 77.5 & 62.7\\
\bottomrule
\end{tabular}
\caption{Same-cache compression controls (\%): truncation and random projection of the cached vectors, with one shared Gaussian matrix and no label fitting.}
\label{tab:retrieval-dimensions}
\end{table}

\begin{table}[t]
\centering\footnotesize\setlength{\tabcolsep}{2pt}
\begin{tabular}{@{}lrrr@{}}
\toprule
Fusion & NDCG@10 & R@10 & Cmpl.@10\\
\midrule
BM25 + native 4096-d & 62.2 & 70.3 & 55.3\\
BM25 + compact 512-d & 65.8 & 74.5 & 60.0\\
BM25 + SkillRouter-Emb. & 65.7 & 73.5 & 58.9\\
BM25 + SkillRet-Emb.-8B & 70.0 & 78.9 & 66.1\\
\bottomrule
\end{tabular}
\caption{Lexical--dense fusion on the SkillRet TEST split (\%). Equal-weight reciprocal rank fusion ($k=60$) over cached Top-100 lists; compact uses seed 17.}
\label{tab:retrieval-fusion}
\end{table}

\paragraph{Retrieval-only comparisons.} BM25 supplies a lexical reference; bge-large, Qwen3-Embedding and harrier supply frozen general-purpose dense references. SkillRouter-Embedding and SkillRet-Embedding use their released skill-tuned checkpoints without further training here. The same-encoder 4096-d and truncated 512-d rows control for encoder identity and dimension; only the compact rows add a learned compressor. Every row ranks skills on the same official test and uses its relevance labels. Completeness@10 requires retrieving every required skill for a query. These comparisons do not run an executor, fit KeyOpt or read execution events.
Table~\ref{tab:skillret} reports the official SkillRet comparison; Tables~\ref{tab:retrieval-dimensions} and~\ref{tab:retrieval-fusion} add compression and fusion controls. The three-seed compact mean improves NDCG@10 by 6.9 points over the 4096-d embeddings of the same frozen encoder and by 8.7 points over truncation, while storing one eighth of the key dimensions. An overlap audit found no exact normalized skill-body or query overlap between TRAIN and TEST, but 518 skill names overlap. For 3,766 test queries, no relevant skill matches TRAIN by name or text.

\begin{table}[t]
\centering\small\setlength{\tabcolsep}{3pt}
\begin{tabular}{@{}lrrrr@{}}
\toprule
Required skills & Queries & 4096-d & 512-d & $\Delta$ pp\\
\midrule
One & 2,143 & 84.2 & 89.7 & +5.5\\
Two & 1,703 & 32.0 & 45.1 & +13.1\\
Three & 546 & 7.0 & 11.3 & +4.3\\
\midrule
Two or three & 2,249 & 25.9 & 36.9 & +11.0\\
\quad Unseen only & 1,838 & 28.7 & 39.9 & +11.2\\
\bottomrule
\end{tabular}
\caption{SkillRet Completeness@10 (\%) by number of relevant skills; unseen: queries without a relevant TRAIN match. Compact 512-d averages three seeds.}
\label{tab:skillret-cardinality}
\end{table}

\paragraph{Coverage by skill count.}
We stratify the saved official scores by the number of annotated relevant skills, without generating new rankings. Of 4,392 queries, 2,143 have one relevant skill, 1,703 have two and 546 have three. In Table~\ref{tab:skillret-cardinality}, Completeness@10 rises by 5.5 points for single-skill queries and 11.0 points for multi-skill queries. For seed 17, the multi-skill group gains 350 complete queries and loses 98, a net gain of 252; seeds 29 and 43 gain 279 and 210. On the 1,838 multi-skill queries whose relevant skills match no TRAIN skill by name or body, coverage rises from 28.7\% to 39.9\%.

We average the three seed scores within each query and bootstrap paired queries 5,000 times (PCG64 seed 20261011, linear percentile intervals). The 95\% interval for the multi-skill gain is [9.3,12.6] points; for its absolute gain minus the single-skill gain, it is [3.5,7.4]. Every skill-count group gains, from 4.3 points for three-skill queries to 13.1 for two-skill queries, with groups starting from different baseline coverage. The gain is therefore one of retrieval coverage: compact keys bring all required skills into the pool more often.

\paragraph{Cached controls.} The controls add nine rankings per query: four equal-weight BM25--dense reciprocal rank fusion (RRF) combinations \citep{cormack2009rrf}, normalized truncation to 128/256/1024/2048 dimensions, and Gaussian random projection to 512 dimensions with seed 17. RRF sums $1/(60+r)$ for one-based ranks $r$ in each frozen Top-100 list; absent candidates contribute zero. The random matrix is shared by queries and skills, sampled with scikit-learn without fitting labels. All projections use the same cached Qwen3-Embedding-8B vectors, normalized again after projection, and FAISS exact cosine search. Fusion retains each released encoder's original document serialization. They ran with two CPU threads (21.0\,s for generation; 14.7\,s for scoring and statistics), with about 1.3\,GiB peak resident memory. These are cached-operation costs, excluding encoding.

\paragraph{Paired retrieval intervals.} The compact key (seed 17) exceeds the 512-d random projection by 10.27 NDCG@10 points (paired-query bootstrap 95\% interval [9.56,10.99]); 2048-d truncation differs from the native 4096-d vectors by 0.16 points ([${-}0.06$,0.38]). Intervals resample the 4,392 test queries 2,000 times with seed 17.

\paragraph{Why the smaller key can retrieve better.} The compressor learns relevance rather than reconstructing all coordinates of the original embedding. Queries keep their fixed 512-d representation, while the document-side correction in Equation~\ref{eq:compact-key} can read all 4,096 coordinates. Training therefore changes which documents align with a request, without changing the document the executor reads. The 512-d truncation and random-projection controls separate this learned alignment from simply using fewer dimensions. The unseen multi-skill results further show that the gain includes bringing all required documents into the pool, rather than only improving the first relevant match. Compression thus yields a representation specialized for retrieval rather than a lossy copy of the embedding.

\label{app:compressor-transfer}
The SkillRet compressor maps 4096 to 512 dimensions using TRAIN revision \nolinkurl{a050ad233a504a43135bafe8cdf45574052b5729} \citep{kang2026skillret}: 63,259 requests, 10,123 skills and 127,190 relevance associations. An internal skill-held-out split chose five epochs; three seeds then trained on all TRAIN. Encoder weights stayed frozen; official tests were unused. Internal R@1 was 66.75\%, versus 53.70\% natively at 512-d.

On the 665 SRA development queries, mean R@8 was 72.52\%, versus 65.60\% for the native 512-d prefix of the same encoder.

\begin{table}[t]
\centering\small\setlength{\tabcolsep}{2.4pt}
\begin{tabular}{@{}lrrrrr@{}}
\toprule
Index & R@1 & R@5 & R@10 & MRR & nDCG\\
\midrule
\multicolumn{6}{@{}l}{\textit{SRA (665 queries, 26K skills)}}\\
\rowcolor{tint}Compact, seed 17 & 44.4 & 64.9 & 73.9 & 60.9 & 60.9\\
Compact, seed 29 & 44.4 & 66.6 & 74.4 & 61.8 & 61.5\\
Compact, seed 43 & 45.3 & 67.4 & 74.9 & 62.5 & 62.5\\
Native 4096-d & 46.1 & 66.8 & 71.8 & 62.5 & 62.1\\
Truncated & 45.0 & 60.4 & 66.8 & 59.7 & 58.4\\
SRA-supervised & 50.2 & 82.7 & 91.8 & 72.4 & 75.7\\
\midrule
\multicolumn{6}{@{}l}{\textit{SRA ToolQA subset (200 queries)}}\\
\rowcolor{tint}Compact, seed 17 & 26.0 & 51.0 & 60.5 & 36.0 & 41.8\\
Compact, seed 29 & 26.0 & 50.0 & 58.0 & 35.6 & 41.0\\
Compact, seed 43 & 29.5 & 53.0 & 61.5 & 39.7 & 44.9\\
Native 4096-d & 30.0 & 38.5 & 43.0 & 33.2 & 35.5\\
Truncated & 28.5 & 32.0 & 35.0 & 30.3 & 31.4\\
SRA-supervised & 51.5 & 78.0 & 84.5 & 62.8 & 68.1\\
\midrule
\multicolumn{6}{@{}l}{\textit{AppWorld, 463 task APIs (57 tasks)}}\\
\rowcolor{tint}Compact, seed 17 & 9.7 & 22.8 & 39.4 & 63.0 & 36.4\\
Compact, seed 29 & 10.8 & 25.4 & 37.3 & 66.9 & 36.7\\
Compact, seed 43 & 12.2 & 27.5 & 37.4 & 71.0 & 38.3\\
Native 4096-d & 8.6 & 22.6 & 36.3 & 61.5 & 34.0\\
Truncated & 9.7 & 20.5 & 34.7 & 60.3 & 33.4\\
SRA-supervised & 10.9 & 24.1 & 33.1 & 65.6 & 33.6\\
\midrule
\multicolumn{6}{@{}l}{\textit{AppWorld, all 469 APIs (57 tasks)}}\\
\rowcolor{tint}Compact, seed 17 & 6.0 & 13.8 & 23.9 & 63.0 & 26.9\\
Compact, seed 29 & 6.6 & 15.4 & 22.7 & 66.9 & 27.0\\
Compact, seed 43 & 7.5 & 16.8 & 22.9 & 71.0 & 28.3\\
Native 4096-d & 5.4 & 14.2 & 22.0 & 61.5 & 25.2\\
Truncated & 6.0 & 12.9 & 21.0 & 60.3 & 24.6\\
SRA-supervised & 6.8 & 14.7 & 20.3 & 65.6 & 24.9\\
\midrule
\multicolumn{6}{@{}l}{\textit{ToolQA (332 tasks, 785 documents)}}\\
Compact 512-d & 49.7 & 79.5 & 89.8 & 63.3 & 69.7\\
SkillRouter bi-encoder & 8.4 & 34.9 & 42.8 & 19.3 & 24.9\\
SkillRouter + reranker & 46.7 & 56.9 & 57.2 & 51.6 & 53.1\\
\bottomrule
\end{tabular}
\caption{Retrieval quality (\%) of compact keys, the embeddings they compress, truncation and a separately supervised compressor; MRR and nDCG at 10.}
\label{tab:retrieval-full}
\end{table}

\begin{figure*}[t]
\centering\includegraphics[width=\textwidth]{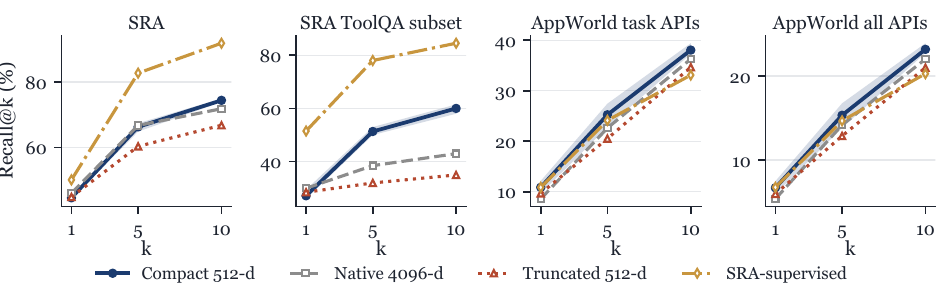}
\caption{Recall@$k$ on four libraries (SRA, its ToolQA subset, two AppWorld API sets): compact keys (three seeds, mean and range), the embeddings they compress, truncation and a separately supervised compressor.}
\label{fig:retrieval-curves}
\end{figure*}
Table~\ref{tab:retrieval-full} and Figure~\ref{fig:retrieval-curves} report all retrieval metrics. The learned projection improves Recall@10 over the 4096-d embedding on each library of Figure~\ref{fig:retrieval-curves}, and the gain is largest on the ToolQA subset of SRA (60.0 versus 43.0\%), where many skills describe the same databases. On the separate 332-task ToolQA execution library, compact keys reach 89.8\% Recall@10. Errors therefore include both missing guides and incorrect ordering among retrieved guides: KeyOpt reranks the existing pool, while event recall can add omitted candidates.

\begin{table}[t]
\centering\small\setlength{\tabcolsep}{3pt}
\begin{tabular}{@{}lrr@{}}
\toprule
Operation (CPU) & Median & P90\\
\midrule
Compact Top-10 & 0.31\,ms & 0.40\,ms\\
Default index and rendering & 0.33\,ms & 0.59\,ms\\
SkillRouter, cached scores & 0.25\,ms & 0.32\,ms\\
GoSkills retrieval and rendering & 172\,ms & 181\,ms\\
\midrule
Storage & \multicolumn{2}{r}{bytes}\\
\midrule
Compact keys (785) & \multicolumn{2}{r}{1,607,680}\\
KeyOpt residual and offsets & \multicolumn{2}{r}{1,612,868}\\
SkillRouter document vectors & \multicolumn{2}{r}{3,215,360}\\
\bottomrule
\end{tabular}
\caption{Selection cost per ToolQA task (332 tasks, three repetitions) and index storage. SkillRouter uses cached encoder and reranker scores.}
\label{tab:efficiency}
\end{table}
Table~\ref{tab:efficiency} measures cached selection cost, excluding encoder, online reranker and reviewer calls. A compact Top-10 query takes 0.31\,ms on two CPU threads; adding KeyOpt and document rendering takes 0.33\,ms. In a cached microbenchmark over 1,000 skills, a 512-d Top-10 takes 0.060\,ms against 0.211\,ms at 4,096 dimensions (3.5$\times$), with 2.0 against 16.4\,MB of base keys. GoSkills' retrieval and contract rendering takes 172\,ms per task. A review call reads 13.1k/11.5k input tokens and takes a median 10.7/5.0 seconds with GPT-5.5/DeepSeek (332 ToolQA reviews each). Base-key storage excludes residuals, the reranker, events and encoder weights. Support acquisition and fitting are one-off costs outside these per-task figures.

\section{ToolQA Results by Query Type, Domain and First Choice}
\label{app:more-results}
\begin{table*}[t]
\centering\footnotesize\setlength{\tabcolsep}{4pt}\renewcommand{\arraystretch}{1.04}
\begin{tabular*}{\textwidth}{@{\extracolsep{\fill}}lcrrrrrrr@{}}
\toprule
 & & \multicolumn{2}{c}{Query type (\%)} & \multicolumn{4}{c}{Database (\%)} & \\
\cmidrule(lr){3-4}\cmidrule(lr){5-8}
Method & Docs & Simple & Complex & Airbnb & Coffee & Flights & Yelp & Hit@1 (\%)\\
\midrule
\multicolumn{9}{@{}l}{\textbf{DeepSeek-V4.1-Flash}}\\
\midrule
No Skills & 0 & 87.0 & 37.1 & 62.7 & 48.9 & 41.4 & \textbf{85.1} & \na\\
Vector Top-4 & 4 & 87.7 & 59.0 & 65.1 & 87.0 & 51.4 & 80.5 & 63.0\\
Vector Top-1 & 1 & 87.0 & 57.3 & 61.4 & 80.4 & 55.7 & 82.8 & 63.0\\
GoSkills & 2.57 & \textbf{90.9} & 43.8 & 67.5 & 53.3 & 58.6 & 82.8 & 44.9\\
SkillRouter & 1 & 85.7 & 52.8 & 68.7 & 70.7 & 51.4 & 78.2 & 46.7\\
BM25 & 1 & 89.6 & 45.5 & 63.9 & 60.9 & 54.3 & 82.8 & 50.3\\
SkillRet-8B & 1 & 87.0 & 55.6 & 60.2 & 76.1 & 55.7 & \textbf{85.1} & 64.5\\
Tool2Vec & 1 & 88.3 & 65.2 & 69.9 & \textbf{88.0} & 58.6 & 82.8 & 63.6\\
RankGPT & 1 & 88.3 & 60.1 & 66.3 & 83.7 & 55.7 & 82.8 & \textbf{78.3}\\
\midrule
\rowcolor{tint}CompSkill & 1 & \textbf{90.9} & \textbf{67.4} & \textbf{71.1} & \textbf{88.0} & \textbf{65.7} & \textbf{85.1} & 74.7\\
\midrule
\multicolumn{9}{@{}l}{\textbf{GPT-5.6-luna}}\\
\midrule
No Skills & 0 & 85.7 & 36.5 & 60.2 & 52.2 & 41.4 & 80.5 & \na\\
Vector Top-4 & 4 & 87.7 & 61.8 & 67.5 & 91.3 & 58.6 & 73.6 & 63.0\\
Vector Top-1 & 1 & 85.1 & 58.4 & 60.2 & 85.9 & 55.7 & 77.0 & 63.0\\
GoSkills & 2.57 & 85.7 & 50.0 & 66.3 & 60.9 & 52.9 & \textbf{83.9} & 44.9\\
SkillRouter & 1 & 83.1 & 54.5 & 57.8 & 76.1 & 54.3 & 79.3 & 46.7\\
BM25 & 1 & 87.7 & 49.4 & 61.4 & 66.3 & 57.1 & 81.6 & 50.3\\
SkillRet-8B & 1 & 88.3 & 50.6 & 57.8 & 82.6 & 44.3 & 81.6 & 64.5\\
Tool2Vec & 1 & 90.3 & 55.6 & \textbf{73.5} & 80.4 & 50.0 & 78.2 & 62.3\\
RankGPT & 1 & \textbf{90.9} & 61.2 & 68.7 & 88.0 & 55.7 & 82.8 & 78.0\\
\midrule
\rowcolor{tint}CompSkill & 1 & 90.3 & \textbf{65.7} & 69.9 & \textbf{92.4} & \textbf{61.4} & 80.5 & \textbf{82.2}\\
\bottomrule
\end{tabular*}
\caption{ToolQA success by query type (154 simple, 178 complex) and database (83 Airbnb, 92 Coffee, 70 Flights, 87 Yelp), and first-choice accuracy (Hit@1: the first supplied document is the labeled guide), for the runs of Table~\ref{tab:main}. Tool2Vec, RankGPT and CompSkill select per executor; the others do not. Bold: column best per executor.}
\label{tab:hit1}
\end{table*}
Table~\ref{tab:hit1} breaks down Table~\ref{tab:main} by query type and database. CompSkill's gains concentrate on complex queries and on Flights, the weakest database for most methods. On simple queries most methods exceed 85\%. With GPT-5.6-luna, CompSkill supplies the labeled guide first on 82.2\% of tasks, the highest rate. With DeepSeek, its executor-specific calibration departs from the labeled guide more often than RankGPT (74.7\% versus 78.3\%), yet CompSkill completes 260 tasks against 243 (20/3 wins/losses, exact $p<0.001$). The labeled guide is a proxy for the useful document; a calibration fitted to the outcomes of one executor can prefer another guide whenever that executor completes the task with it more reliably on support tasks.

\section{Held-out and Further Evaluations}
\label{app:post}
This appendix reports the held-out ToolQA evaluations and two further benchmarks. GPT-5.5 was withdrawn from our access on 2026-10-14, so GPT-5.6-luna (low reasoning) executes the held-out ToolQA tasks with the unchanged GPT-5.5 fits. DeepSeek replaced the September checkpoint behind \texttt{deepseek-flash} with V4.1-Flash-1017 on 2026-10-17 and served another checkpoint from 2026-10-28; the September checkpoint is no longer served. ToolQA runs keep the fits learned from September support, while the further benchmarks (Appendix~\ref{app:fresh}) use V4.1-Flash-1017 with support collected only from that same checkpoint.

\subsection{Held-out ToolQA Evaluation}
\label{app:heldout}
\paragraph{Cohorts.} The 180 tasks are all SRA ToolQA instances for the eight guides of Table~\ref{tab:guides} that belong to neither the 332 evaluation tasks nor the 318 support tasks. Eighty of them appeared in an earlier exploratory DeepSeek study, so the main DeepSeek cohort excludes them ($n=100$). With the September checkpoint, CompSkill completes 70 of these 100 tasks, level with four-document Vector Top-4 (70) and ahead of Vector Top-1 (64) and SkillRouter (61; 10/1 wins/losses, exact $p=0.012$), and 130 of all 180 against 125, 115 and 110 (18/3 and 22/2 over Vector Top-1 and SkillRouter, exact $p\le0.002$). The 332 new tasks were generated by the original easy and hard notebooks of the official ToolQA repository at a fixed commit, with the same 82 templates and the per-template counts of the 332 evaluation tasks, a fixed seed, and exclusion of every question that duplicates a support, evaluation, held-out or published ToolQA question; five Flights template cells whose question and answer computation disagreed were corrected. Executions that failed with transport errors were retried once.
\paragraph{Inputs.} Queries were encoded with the paper's encoder and instruction, ranked by the frozen compact keys and scored by the same reranker, with the calibrated index, KeyOpt parameters (Appendix~\ref{app:cv-strengths}), events, reviewer prompt and baselines of the main ToolQA experiments, unchanged. The GPT reviewer prompt names GPT-5.5, whose support runs form the events; GPT-5.6-luna answers the reviews, except that 109 evaluation tasks reuse GPT-5.5 reviews of identical prompts.
\paragraph{Comparisons.} On the 180 held-out tasks, CompSkill and the KeyOpt index are compared with each of BM25, SkillRet-8B, Tool2Vec and RankGPT (Holm over eight). On the 332 new tasks, CompSkill is compared with the same four retrievers (Holm over four), and the KeyOpt index with offsets alone, the recall pool with the index, and review with recall (Holm over three). The same families were also run on the 332 evaluation tasks with GPT-5.6-luna, where CompSkill completes 256 against 249 for RankGPT, 238 for Tool2Vec, 226 for SkillRet-8B and 223 for BM25, and with DeepSeek-V4.1-Flash, where it completes 260 against 252, 243, 233 and 219 for Tool2Vec, RankGPT, SkillRet-8B and BM25 (Holm-corrected $p=0.115$ for Tool2Vec and $p\le0.001$ for the others). Table~\ref{tab:heldout-strong-tests} lists every comparison of these families. On the 180 held-out tasks CompSkill completes 124, against 126 for four-document Vector Top-4, 125 for Tool2Vec, 119 for the KeyOpt index, 118 for RankGPT, 106 for Vector Top-1, SkillRouter and BM25, and 105 for SkillRet-8B. It exceeds Vector Top-1 and SkillRouter (22/4 and 23/5 wins/losses, exact $p<0.001$) and BM25 and SkillRet-8B (23/5 and 23/4, Holm $p\le0.006$); its differences from Tool2Vec, RankGPT and four-document Vector Top-4 are not significant.

\paragraph{DeepSeek checkpoints.} The DeepSeek checkpoint behind our support fits was retired during the study (fingerprint \texttt{aeb56401}); later ToolQA runs use the checkpoint served from 2026-10-28 (\texttt{bc19ba97}). On the 180 held-out tasks both checkpoints executed identical inputs for the KeyOpt index and for CompSkill. Outcomes agree on 154 of 177 and 156 of 178 jointly known tasks; the disagreements are symmetric (10/13 and 11/11, exact $p=0.68$ and $1.00$), and totals are 128 versus 132 and 130 versus 131. DeepSeek conditions whose selections differ from earlier executions therefore run on the later checkpoint. With that checkpoint, CompSkill completes 131 held-out tasks, the KeyOpt index 132, offsets alone 129, Tool2Vec 130 (6 unknown) and RankGPT 122; CompSkill against Tool2Vec is 6/5 in wins/losses (Holm $p=1$) and against RankGPT 15/6 ($p=0.157$). On the 332 evaluation tasks, the four stronger baselines reuse identical earlier DeepSeek executions where they exist and otherwise run on the later checkpoint. Outcomes that could not be scored, mostly from output-limit exhaustion (31 baseline and six CompSkill outcomes), were completed with V4.1-Flash-1017, reusing identical earlier executions where they existed.

\paragraph{Cost.} Per GPT-5.6-luna task, CompSkill supplies 1.1k guidance tokens against 5.1k for Vector Top-4. Including the review call, total input per task is 62.0k versus 60.0k tokens with GPT-5.6-luna and 30.3k versus 34.5k with DeepSeek.

\subsection{KeyOpt Fitting Details}
\label{app:cv-strengths}
KeyOpt scores a candidate with Equation~\ref{eq:score}, where $r_t(s)$ is the frozen reranker score, $\beta_s$ the reliability offset, the clipped term the conditional residual of Section~\ref{sec:keyopt}, and $c_t(s)$ the event reading of Equation~\ref{eq:reading}. Offsets are fitted first with the pairwise loss; residuals are then fitted on top of them with a keep penalty that holds every residual near zero on all support queries and lets it move only where matched comparisons disagree with the offsets. Five-fold source-group cross-validation on support tasks alone selects the reading weight $\alpha$, the score cap $c$, the keep weight and whether the residual is used. On ToolQA the selection kept the residual for both executors ($c=0.3$, $\alpha=0$ for GPT; $c=0.2$, $\alpha=0.05$ for DeepSeek). On the 332 evaluation tasks, executed with GPT-5.6-luna from the frozen GPT-5.5 fits, the KeyOpt index completes 254 tasks against 236 with offsets alone (26/8 wins/losses; held-out: Table~\ref{tab:heldout}).

\subsection{Which Errors Does Calibration Fix?}
\label{app:anatomy}
We classify each supplied document as the labeled guide, a \emph{sibling} (another guide for the same database), another benchmark guide, or an unrelated library document, and compare outcomes within the same task (Table~\ref{tab:anatomy}). On the original ToolQA databases a sibling costs 17--25 points against the labeled guide, more than an unrelated document (8--16 points): a near-miss procedure costs more than an unrelated one, and in direct within-task comparisons the difference is positive in all four cells. Siblings are also the errors that description similarity cannot resolve, because both guides describe the same database. The KeyOpt index cuts these errors: with the GPT fit it supplies a sibling on 16 of the 180 held-out tasks versus 37 for Vector Top-1 (exact McNemar $p=0.001$; with the DeepSeek fit 35 versus 37), and on the three further ToolQA databases below it supplies fewer siblings than the uncalibrated index with both executors (79 vs.\ 113, $p=0.002$; 85 vs.\ 113, $p=0.02$), partly by switching to unrelated documents (Appendix~\ref{app:fresh}).

\begin{table}[t]
\centering\small\setlength{\tabcolsep}{4pt}
\begin{tabular}{@{}lcr@{}}
\toprule
Supplied documents & Docs & Successes\\
\midrule
Labeled guide & 1 & 72\\
Labeled + sibling guide & 2 & 67\\
Labeled + other-database guide & 2 & \textbf{75}\\
\bottomrule
\end{tabular}
\caption{Adding a second guide on 96 held-out ToolQA tasks with GPT-5.6-luna (12 per labeled guide). The other-database guide has the sibling's type; both two-document conditions keep each task's document order.}
\label{tab:conflict}
\end{table}

\paragraph{Adding a sibling.} The costs above come from replacing the labeled guide. A controlled run tests adding a guide instead: on 96 of the 180 held-out tasks, GPT-5.6-luna received the labeled guide alone, together with its sibling, or together with the guide of the same type for another database, each once and rendered as for Vector Top-4 (Table~\ref{tab:conflict}). With the sibling it completes 67 tasks, against 75 with the other database's guide (10/2 wins/losses, exact $p=0.039$; difference 8.3 points, bootstrap 95\% interval [2.1, 15.6]), and 72 with the labeled guide alone. The gap holds whether the labeled guide comes first or second (5/1 each). A second document thus costs success when it describes another procedure for the same database, the partial conflict that induces procedural hallucinations \citep{jiang2026conflict}, and supplying one well-chosen guide avoids this conflict altogether.

The labels only proxy usefulness. On BIRD-Skills the gold and sibling table guides are equally useful ($+0.02$ in both executors), and the KeyOpt index moves many choices from gold to sibling tables (182 vs.\ 59 siblings for the uncalibrated index) with little change in success (276 vs.\ 280 and 328 vs.\ 326; Table~\ref{tab:bird}). Calibration follows observed utility for the executor, not annotated relevance.

\subsection{Further Benchmarks and the Applicability Boundary}
\label{app:fresh}
We also evaluated two further benchmarks, each with its own support split and executors' own fits, comparing the single-document calibrated index (without review) with the strongest single-document baseline. The baselines include three retrievers that use the same support feedback or an executor call: Tool2Vec builds document vectors from successful support queries \citep{moon2024tool2vec}, ReMe injects memories extracted from the same support trajectories \citep{cao2026reme}, and RankGPT lets the executor rank the Top-10 \citep{sun2023rankgpt}; BM25 and the SkillRet-8B retriever \citep{kang2026skillret} complete the set. Both benchmarks lack the property behind CompSkill's ToolQA gains: which single document the executor receives barely changes success. Accordingly, the calibrated index stays statistically level with the strongest single-document baseline on both.

\paragraph{Three further ToolQA databases.} SciREX, Agenda and DBLP each have two SRA guides written by the benchmark authors; adding these six guides gives a 791-document library. Of 600 tasks, 283 form the support split and 317 the evaluation split. Support yields only 33 (GPT-5.6-luna) and 31 (DeepSeek) tasks whose two executed guides differ in outcome, against 72 and 87 on the original databases, and the calibrated index often moves to unrelated documents (133 and 145 of 317 choices, versus 47 for the uncalibrated index), a sign that the corrections overfit the sparse support. Even the labeled guide adds only 20 and 37 tasks over No Skills, and on SciREX it falls below No Skills (4 and 7 versus 10 and 8 of 104), where tool observations are truncated and tables are replaced by placeholders (Table~\ref{tab:newdb}). The calibrated index stays within the noise of SkillRet-8B (73 vs.\ 82 and 82 vs.\ 84; $p=0.11$ and $0.85$). It does supply one document and still exceeds four-document Vector Top-4 (24/8 and 26/9 wins/losses, $p<0.01$), and adding event recall without review hurts (18/4 and 21/6, Holm-corrected $p<0.05$), which is why recall and review are used together in CompSkill.

\paragraph{BIRD-Skills.} We built a skill benchmark from the BIRD development set \citep{li2023bird}: 75 table guides generated by a fixed template from the public column descriptions of 11 databases; 621 support, 622 evaluation and 291 untouched reserve questions by a fixed hash. The executor sees the schema DDL and a read-only SQL tool but not BIRD's per-question evidence hints, and the official execution-accuracy scorer decides success. Guides help (No Skills 233 and 308 versus 271--285 and 320--332 for the single-document selectors), but all 13 single-document selectors lie within 14 tasks of each other, and cross-validation set the event weight to zero for both executors (Table~\ref{tab:bird}). The calibrated index does not differ significantly from the strongest baseline (276 vs.\ 285 Tool2Vec; 328 vs.\ 332 SkillRet-8B); the multi-document GoSkills, which supplies several table guides at once, leads with DeepSeek (349, $p=0.02$).

\paragraph{Reading the boundary.} CompSkill helps where confusable guides lead to different execution outcomes and matched support exposes those differences, as on the original ToolQA databases. Where the right guide adds little over an arbitrary one, or support contains few discordant comparisons, there is little for calibration to learn, and it should match rather than exceed strong retrievers.

\begin{table*}[t]
\centering\small
\begin{tabular}{@{}llrr@{}}
\toprule
Tasks & Comparison & W/L & Holm $p$\\
\midrule
Held-out 180 & CompSkill vs.\ SkillRet-8B & 23/4 & 0.002\\
Held-out 180 & CompSkill vs.\ Tool2Vec & 8/9 & 1.000\\
Held-out 180 & CompSkill vs.\ RankGPT & 10/4 & 0.718\\
Held-out 180 & CompSkill vs.\ BM25 & 23/5 & 0.006\\
Held-out 180 & KeyOpt index vs.\ SkillRet-8B & 22/8 & 0.097\\
Held-out 180 & KeyOpt index vs.\ Tool2Vec & 6/12 & 0.718\\
Held-out 180 & KeyOpt index vs.\ RankGPT & 8/7 & 1.000\\
Held-out 180 & KeyOpt index vs.\ BM25 & 24/11 & 0.205\\
\midrule
New 332 & CompSkill vs.\ BM25 & 49/7 & $<$0.0001\\
New 332 & CompSkill vs.\ SkillRet-8B & 37/3 & $<$0.0001\\
New 332 & CompSkill vs.\ Tool2Vec & 21/6 & 0.012\\
New 332 & CompSkill vs.\ RankGPT & 13/2 & 0.012\\
\midrule
New 332 & KeyOpt index vs.\ Offsets only & 12/4 & 0.230\\
New 332 & 6+4 pool vs.\ KeyOpt index & 9/4 & 0.292\\
New 332 & CompSkill vs.\ 6+4 pool & 9/3 & 0.292\\
\midrule
Development 332 & CompSkill vs.\ BM25 & 44/11 & $<$0.0001\\
Development 332 & CompSkill vs.\ SkillRet-8B & 36/6 & $<$0.0001\\
Development 332 & CompSkill vs.\ Tool2Vec & 27/9 & 0.008\\
Development 332 & CompSkill vs.\ RankGPT & 12/5 & 0.143\\
\midrule
Development 332 & KeyOpt index vs.\ Offsets only & 26/8 & 0.009\\
Development 332 & 6+4 pool vs.\ KeyOpt index & 2/3 & 1.000\\
Development 332 & CompSkill vs.\ 6+4 pool & 12/9 & 1.000\\
\bottomrule
\end{tabular}
\caption{Paired comparisons with GPT-5.6-luna, one block per Holm family (wins/losses, exact two-sided McNemar on jointly known pairs). The held-out family contains eight tests: four for CompSkill and four for the KeyOpt index. Development 332: the 332 evaluation tasks; 6+4 pool: KeyOpt index on six semantic and four recalled candidates.}
\label{tab:heldout-strong-tests}
\end{table*}

\begin{table*}[t]
\centering\small\setlength{\tabcolsep}{4pt}
\begin{tabular}{@{}llccc@{}}
\toprule
Cohort & Executor & Label $-$ sibling & Label $-$ unrelated & Unrelated $-$ sibling\\
\midrule
ToolQA 332 (dev.) & GPT & $+0.24$ [$+0.17$, $+0.31$] & $+0.11$ [$+0.07$, $+0.16$] & $+0.09$ [$+0.01$, $+0.17$]\\
 & DeepSeek & $+0.17$ [$+0.10$, $+0.24$] & $+0.08$ [$+0.02$, $+0.13$] & $+0.01$ [$-0.04$, $+0.08$]\\
ToolQA held-out 180 & GPT & $+0.25$ [$+0.13$, $+0.36$] & $+0.09$ [$0.00$, $+0.18$] & $+0.11$ [$0.00$, $+0.26$]\\
 & DeepSeek & $+0.25$ [$+0.13$, $+0.37$] & $+0.16$ [$+0.04$, $+0.28$] & $+0.09$ [$0.00$, $+0.20$]\\
ToolQA, three new databases & GPT & $+0.04$ [$+0.02$, $+0.06$] & $+0.09$ [$+0.05$, $+0.12$] & $+0.02$ [$0.00$, $+0.04$]\\
 & DeepSeek & $+0.06$ [$+0.02$, $+0.10$] & $+0.11$ [$+0.06$, $+0.15$] & $+0.03$ [$-0.01$, $+0.08$]\\
BIRD-Skills & GPT & $+0.02$ [$-0.02$, $+0.06$] & $+0.06$ [$+0.02$, $+0.10$] & $-0.04$ [$-0.10$, $+0.01$]\\
 & DeepSeek & $+0.02$ [$-0.02$, $+0.05$] & $+0.03$ [$-0.01$, $+0.07$] & $-0.03$ [$-0.09$, $+0.03$]\\
\bottomrule
\end{tabular}
\caption{Within-task success differences between supplied-document categories (95\% intervals from 10,000 task resamples). GPT is GPT-5.5 on the 332 evaluation tasks and GPT-5.6-luna on the three other cohorts. Sibling: a guide of the same database (ToolQA) or a non-gold table of the same database (BIRD). Each column uses only the tasks on which both of its compared conditions were executed with the same executor.}
\label{tab:anatomy}
\end{table*}

\begin{table}[h]
\centering\small\setlength{\tabcolsep}{4pt}
\begin{tabular}{@{}lrr@{}}
\toprule
Method ($n=317$) & GPT-5.6 & DeepSeek-1017\\
\midrule
No Skills & 64 & 49\\
Vector Top-1 & 59 & 62\\
SkillRouter & 61 & 58\\
BM25 & 59 & 57\\
SkillRet-8B & 82 & 84\\
Tool2Vec & 79 & 78\\
RankGPT & 72 & 73\\
ReMe & 67 & 73\\
\midrule
Vector Top-4 (4 docs) & 57 & 65\\
GoSkills (group) & 76 & 74\\
\midrule
Compact keys + reranker & 70 & 64\\
Calibrated index (offsets) & 74 & 82\\
Calibrated index (CV) & 73 & 82\\
\quad without event reading & 72 & 82\\
\quad flat memory instead of events & 71 & 76\\
\quad + event recall, no review & 59 & 67\\
\midrule
Labeled guide (oracle) & 84 & 86\\
\bottomrule
\end{tabular}
\caption{Three further ToolQA databases (SciREX, Agenda, DBLP; 317 evaluation tasks, 791 documents). Unknowns (at most one per cell) count as failures.}
\label{tab:newdb}
\end{table}

\begin{table}[h]
\centering\small\setlength{\tabcolsep}{4pt}
\begin{tabular}{@{}lrr@{}}
\toprule
Method ($n=622$) & GPT-5.6 & DeepSeek-1017\\
\midrule
No Skills & 233 & 308\\
Vector Top-1 & 275 & 325\\
SkillRouter & 275 & 325\\
BM25 & 273 & 322\\
SkillRet-8B & 271 & 332\\
Tool2Vec & 285 & 331\\
RankGPT & 272 & 329\\
ReMe & 276 & 320\\
\midrule
Vector Top-4 (4 docs) & 292 & 327\\
GoSkills (group) & 280 & 349\\
\midrule
Compact keys + reranker & 280 & 326\\
Calibrated index (offsets) & 285 & 332\\
Calibrated index (CV) & 276 & 328\\
\quad without event reading & 276 & 328\\
\quad flat memory instead of events & 276 & 328\\
\quad + event recall, no review & 278 & 328\\
\bottomrule
\end{tabular}
\caption{BIRD-Skills (622 evaluation tasks, no evidence hints, 860 documents). Unknown outcomes (at most one per cell) count as failures.}
\label{tab:bird}
\end{table}

\section{Extended Related Work}
\label{app:rw}
\paragraph{Skill ecosystems.} Agent skills package procedures as documents with optional scripts \citep{anthropic2025skills,agentskills2026arch}. Surveys organize their lifecycle, architecture and risks \citep{jiang2026sok,badhe2026survey,li2026dynamicsurvey}, and large-scale studies catalog public repositories and their reusability \citep{li2026organizing,wang2026skillcorpus,zhang2026reusable}. Benchmarks measure whether skills help at all \citep{li2026skillsbench,wang2026notall,tank2026regression,jiang2026demystifying} and whether agents actually follow them \citep{cho2026following}; their common finding, that a relevant skill can hurt, motivates CompSkill.
\paragraph{Retrieval and routing.} Beyond dense encoders \citep{karpukhin2020dpr,zhang2025qwen3embedding} and tool retrievers \citep{qin2024toolllm,qu2024colt,shi2025toolret,liu2024toolnet,moon2024tool2vec}, skill retrievers add multi-stage pipelines \citep{zhou2025skillflow}, bias calibration \citep{xiao2026skillsight}, pseudo-queries \citep{ding2026skill2query}, neighbor contrast \citep{wang2026neighbors}, reasoning \citep{jiang2026skillreason}, prospective queries \citep{liu2026imagine}, task decomposition \citep{chen2026taskdecomp}, diversity \citep{wei2026beyondtopk} and marketplace-scale indexing \citep{yang2026skillseek,kolluru2026comparative}. Routing studies elicit native routing from frozen models \citep{chen2026routerwithin}, compose skills \citep{gao2026compositional} and study router-aware skill-injection attacks \citep{najjar2026surviving}. Graph methods connect skills by dependency, causality or co-use \citep{li2026gos,zeng2026goskills,xia2026grasp,li2026caskg,yao2026skilltrace,skillgraph2026,fu2026segos}. Most rank by what a skill describes; SE-GoS and SkillGraph further evolve their skill graphs from execution trajectories or reinforcement-learning feedback \citep{fu2026segos,skillgraph2026}. CompSkill writes matched same-executor outcomes into keys and keeps attributed events for cited review.
\paragraph{Learning from execution.} Agents improve by storing reflections, insights, workflows and procedures \citep{shinn2023reflexion,zhao2024expel,wang2024awm,cao2026reme,suzgun2025cheatsheet,zhang2025ace,ouyang2025reasoningbank} or by inducing and revising skills \citep{zheng2025skillweaver,wang2025asi,yang2026skillopt,skillos2026,skill12026,he2026skillcommit,zheng2026skillprox,zhang2026memskill,wang2026skillx}. Closest to us, paired execution audits revise skills \citep{gao2026selfevoskill}, replayable evidence grounds evolution \citep{zhou2026eviskill}, reliability is tracked across coalitions \citep{zhao2026coalition}, and activation is gated by counterfactual or executability evidence \citep{guo2026skillapt,ashkenazi2026gating,he2026radeg}. CompSkill differs by writing evidence into retrieval keys and cited events while leaving documents unchanged, so every edit stays local, bounded and auditable \citep{geva2021keyvalue,meng2022rome}. Graph memories \citep{edge2024graphrag,gutierrez2024hipporag,xu2026hyperskill} motivate our attributed hyperedges with provenance.
\paragraph{Learning to rank from outcomes.} KeyOpt is a pairwise learning-to-rank model \citep{joachims2002optimizing,burges2005ranknet} whose preferences come from matched executions rather than clicks or labels. Comparing two documents on the same task controls their shared task context; an additive task-ease term cancels under the explicit model in Proposition~\ref{prop:identify}. As with learning from logged clicks \citep{joachims2017unbiased}, the guarantee rests on an explicit model of the observation process. NUDGE also makes bounded edits to individual embeddings \citep{zeighami2025nudge}; KeyOpt differs in its supervision, matched execution outcomes rather than relevance labels, and in keeping the events behind each edit, so a round's changes can be traced to the recorded comparisons behind them.
\paragraph{Models as rerankers and judges.} Listwise reranking \citep{sun2023rankgpt} and LLM-as-judge evaluation \citep{zheng2023judging} show that a language model can order candidates from their text. CompSkill's reviewer differs in three ways: it is the executor's own model, it receives attributed outcomes in addition to documents, and it must justify its choice with a verbatim quote and event identifiers, which makes every decision checkable. Because quotes and identifiers are validated mechanically, an invalid review never changes the selection; it falls back to the calibrated proposal.

\section{Future Work}
\label{app:future}
Several directions follow. First, KeyOpt is fitted offline from one support set; running rounds online on fresh tasks would test the mechanism under distribution shift. Second, events are shared only between versions of one executor (GPT-5.5 and GPT-5.6-luna on ToolQA); sharing them across model families could help small models with sparse support. Third, keys could be combined with skill rewriting \citep{yang2026skillopt,zheng2026skillprox} with the original document as a fallback.

\paragraph{Operation-level evidence.} Events could additionally be indexed by the tool operations they exercised, such as cumulative filtering or extreme-value ordering. Reading and review could then use both request similarity and the recorded procedure when selecting evidence for a new request.

\paragraph{Conflict-aware selection.} Context that partially conflicts with a selected skill induces procedural hallucinations even after the right skill is chosen \citep{jiang2026conflict}, and a sibling guide is exactly such a near-miss. Review could check the request against each candidate's scope statement and reject candidates whose procedure conflicts with it, and multi-document selectors could be compared by whether their sets mix conflicting procedures.

\end{document}